\PassOptionsToPackage{table,dvipsnames}{xcolor}
\pdfoutput=1

\documentclass[11pt]{article}

\usepackage{ACL2023}

\usepackage{times}
\usepackage{latexsym}

\usepackage[T1]{fontenc}
\usepackage[utf8]{inputenc}

\usepackage{microtype}

\usepackage{inconsolata}
\usepackage{multirow}
\usepackage{booktabs}
\usepackage{xspace}
\usepackage{graphicx}
\usepackage{float}
\usepackage{multirow}
\usepackage{array}
\usepackage{bm}
\usepackage{amssymb}
\usepackage[most]{tcolorbox}  % 加载宏包
\newtcolorbox{problemBox}{
  colback=gray!10,    % 背景色（10% 灰）
  colframe=gray!60,   % 边框色（80% 灰）
  coltitle=black,     % 标题颜色
  fonttitle=\bfseries, % 标题字体
  boxrule=0.5pt,      % 边框线宽
  arc=3pt,            % 圆角
  left=5pt, right=5pt, top=3pt, bottom=3pt, % 内边距
  width=0.48\textwidth
}

\usepackage{hyperref, enumitem}
\setitemize[1]{itemsep=0pt,partopsep=0pt,parsep=0pt,topsep=0pt,leftmargin=12pt}

\usepackage{pifont}
\usepackage[utf8]{inputenc}

\newcommand{\cmark}{\textcolor{green!70!black}{\ding{51}}}
\newcommand{\xmark}{\textcolor{red}{\ding{55}}}

\usepackage{enumitem}
\usepackage{hyperref}

\newcommand{\benchName}{\textsc{EssayBench}\xspace}

\title{\benchName: Evaluating Large Language Models in Multi-Genre Chinese Essay Writing}

\author{
\textbf{Fan~Gao\textsuperscript{1}},
\textbf{Dongyuan~Li\textsuperscript{1}},
\textbf{Ding~Xia\textsuperscript{1}},
\textbf{Fei~Mi\textsuperscript{2}},\\
\textbf{Yasheng~Wang\textsuperscript{2}},
\textbf{Lifeng~Shang\textsuperscript{2}},
\textbf{Baojun~Wang\textsuperscript{2}}\\
\textsuperscript{1}The University of Tokyo,
\textsuperscript{2}Huawei Noah’s Ark Lab\\
\small\texttt{fangao0802@gmail.com}
\small\texttt{puking.w@huawei.com}
  }

\begin{document}
\maketitle

\begin{abstract}
\text{Chinese essay} writing and its evaluation are critical in educational contexts, yet the capabilities of Large Language Models (LLMs) in this domain remain largely underexplored. Existing benchmarks often rely on coarse-grained text quality metrics, largely overlooking the structural and rhetorical complexities of Chinese essays, particularly across diverse genres. To address this gap, we propose \benchName, a multi-genre benchmark specifically designed for Chinese essay writing across four major genres: \textit{Argumentative}, \textit{Narrative}, \textit{Descriptive}, and \textit{Expository}. We curate and refine a total of 728 real-world prompts to ensure authenticity and meticulously categorize them into the \textit{Open-Ended} and \textit{Constrained} sets to capture diverse writing scenarios. To reliably evaluate generated essays, we develop a fine-grained, genre-specific scoring framework that hierarchically aggregates scores. We further validate our evaluation protocol through a comprehensive human agreement study. Finally, we benchmark 15 large-sized LLMs, analyzing their strengths and limitations across genres and instruction types. With \benchName\footnote{\hyperlink{dataset and code will be publicly released.}{https://anonymous.4open.science/r/EssayBench-2B14}}, we aim to advance LLM-based Chinese essay evaluation and inspire future research on improving essay generation in educational settings.

% While current Large Language Models show strong abilities in generating text, their abilities in essay writings has not been comprehensively explored and evaluated. Existing bench evaluated coarse-grained dimension of the generated text quality, rather view it in a essay background. Furthermore, the Chinese essays typed in multi-genres with complex lexical and discourse features. To over the gap of this, we propose \benchName, a fined-grained benchmark tailored for Chinese essay writings in multi-genres (Argumentative, Narrative, Descriptive and Expository). To ensure the authenticity and diversity, we collect and refine totally 728 prompts in the real-world scenarios for evaluation. To evaluate the essays reliably, we propose a fine-grained genre-oriented evaluation framework and aggregate the score in a dependency-aware way. To evaluate the effectiveness and robustness of our evaluation framework, we conduct a comprehensive human agreement study. To understand how LLMs performs in writing different genres of essays, we benchmark 16 large-size LLMs for comparing and analysis their strengths and weakness in different genres and instruction types. We then highlight the weaknesses of LLMs in essay writing and point towards potential avenues for future work.
\end{abstract}

\begin{table*}[t]
\small
\centering
\renewcommand{\arraystretch}{1.2}
\setlength{\tabcolsep}{3.6pt}  % 调整列间距
\begin{tabular}{lc|ccc|ccc}
\toprule
\multirow{2}{*}{\textbf{Benchmark}} & \multirow{2}{*}{\textbf{Num.}} & \multicolumn{3}{c|}{\textbf{Dataset Composition}} & \multicolumn{3}{c}{\textbf{Evaluation Method}} \\
 \ & \ & Domain & Language & Constraints & LLM & F.G. Traits & Scoring \\
\midrule
C-Eval~\cite{huang2023ceval} & 13,948 & General Tasks & ZH & \xmark & \xmark & \xmark & - \\
AlignBench~\cite{liu-etal-2024-alignbench} & 683 & General Tasks & ZH & \xmark & \cmark & \xmark & Direct \\
LongBench-Write~\cite{bai2024longwriter} & 120 &  General Writing & ZH\&EN & \xmark & \cmark & \xmark & Direct \\
HelloBench~\cite{que2024hellobench} & 647 & General Tasks & EN & \xmark & \cmark & \xmark & Weighted\\
WritingBench~\cite{wu2025writingbench} & 1239 & General Writing & ZH\&EN & \cmark & \cmark & \xmark & Direct\\
\midrule
\textbf{\benchName (Ours)} & \textbf{728} & \textbf{Essay Writing} & \textbf{ZH} & \cmark & \cmark & \cmark & \textbf{Weighted} \\
\bottomrule
\end{tabular}
\caption{Comparison of \benchName with other benchmarks in terms of size, composition, and evaluations.}
\label{tab:benchmark_comparison}
\end{table*}

\begin{figure}
    \centering
    \includegraphics[width=\linewidth]{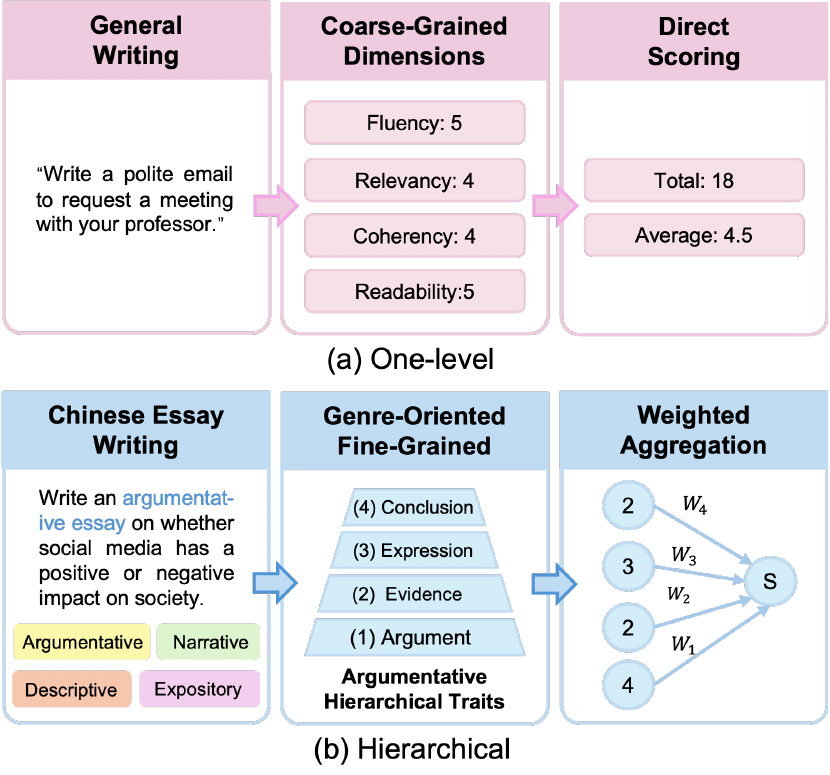}
    \caption{Comparison between coarse-grained evaluation methods (a) and our fine-grained and genre-oriented framework for \benchName(b).}
    \label{fig:comparison}
\end{figure}

\section{Introduction}

Large Language Models (LLMs)~\cite{brown2020language, touvron2023llama,qwen,team2023internlm} have achieved impressive results in text generation, with growing applications in education, including automated writing support and feedback~\cite{gao2024evaluating}. 
Among these tasks, essay writing plays a central role in language learning and assessment~\cite{venkatraman2025collabstorymultillmcollaborativestory, miura2025understandingsupportingformalemail, wen2025interactivesurveyllmbasedpersonalizedinteractive}. However, the lack of robust evaluation frameworks for generated essays limits the development and deployment of LLMs in real-world educational settings~\cite{kim-etal-2025-biggen}.

% Large language models (LLMs) ~\cite{brown2020language, touvron2023llama, team2023internlm, qwen} have demonstrated a remarkable ability to produce realistic, and contextually appropriate texts, and in some cases surpass human writing in terms of surface-level understandability and naturalness~\cite{10.1145/3591196.3596612, radivojevic2024humanperceptionllmgeneratedtext, abassy2025llmdetectaivetoolfinegrainedmachinegenerated}. Yet, despite these successes, accurately evaluating the quality of the generated texts remains a persistent challenge, which is crucial for pinpointing limitations and identifying areas for improvement in LLMs~\cite{hu2024unveilingllmevaluationfocused, kim-etal-2025-biggen}.

% a fundamental question remains largely unsolved: \textit{How can we reliably assess the quality of the generated texts in a way that truly reflects human preferences?}

 % Their capacity to follow real-world complex instructions makes them applicable to a wide range of tasks, from drafting emails and summarizing documents to creative storytelling and survey generation~\cite{kumar2024longlampbenchmarkpersonalizedlongform, venkatraman2025collabstorymultillmcollaborativestory, miura2025understandingsupportingformalemail, wen2025interactivesurveyllmbasedpersonalizedinteractive}. 

As shown in \autoref{fig:comparison}, current predominant LLM-as-a-judge strategies~\cite{zheng2023judgingllmasajudgemtbenchchatbot, li2025generationjudgmentopportunitieschallenges} for assessing texts mainly fall into two paradigms. One relies on meta-evaluation to judge response quality in terms of fluency, relevancy, coherency, readability, and hallucination~\cite{liu-etal-2023-g, chen2023exploringuselargelanguage, hashemi-etal-2024-llm, fu-etal-2024-qgeval}, while the other employs downstream tasks (e.g., question-answering) as proxies for measuring informational richness and accuracy~\cite{tan2024proxyqa, que2024hellobench, lee2025checkevalreliablellmasajudgeframework}. Although these methods yield valuable insights, they exhibit two fundamental weaknesses. First, the evaluation criteria remain overly coarse-grained, i.e., current LLMs consistently achieve high scores in fluency, relevancy, and coherency~\cite{gu2025surveyllmasajudge}, making it difficult to reveal failure modes or specific weaknesses. 
Second, existing evaluation methods fail to capture the unique characteristics of essays like logographic characters, complex constructions, and rhetorical traditions, although several benchmarks like \textit{AlignBench}~\cite{liu-etal-2024-alignbench} and \textit{WritingBench}~\cite{wu2025writingbench} have turned attention to evaluating general Chinese writing.

% Second, existing evaluation methods face significant limitations in Chinese essays. Although several benchmarks like \textit{AlignBench}~\cite{liu-etal-2024-alignbench} and \textit{WritingBench}~\cite{wu2025writingbench} have turned attention to evaluating general Chinese writing, they fail to capture the unique characteristics of essays like logographic characters, complex constructions, and rich rhetorical traditions.

Moreover, Chinese literary and expository practices differ markedly across genres: argumentative essays demand logical structure and persuasive rhetoric~\cite{wachsmuth-etal-2017-computational}; narratives require compelling plot development and character voice~\cite{somasundaran-etal-2018-towards}; descriptive writings emphasize vivid imagery and sensory detail~\cite{mccarthy1998descriptive}; and expository passages call for clarity, organization and factual precision~\cite{balepur-etal-2023-expository}. However, existing evaluation frameworks largely overlook genre-specific criteria, limiting their ability to reflect the nuanced demands of Chinese essay writing. This motivates our central research question as follows:
%These genre-specific requirements highlight the inadequacy of existing evaluation frameworks. We therefore address the following research question: 
%\textbf{How can we reliably assess the quality of LLM-generated Chinese essays in ways that truly reflect genre-specific conventions?}
% \vspace{-3pt}
\begin{center}
\begin{problemBox}
\textit{How can we reliably assess the quality of LLM-generated Chinese essays in ways that truly reflect genre-specific conventions?}
\end{problemBox}    
\end{center}
% \vspace{-3pt}

In this paper, we introduce \benchName, a \textbf{fine-grained} and \textbf{multi-genre} benchmark tailored for Chinese essay writing. \benchName covers four widely recognized genres in Chinese education: \textit{Argumentative}, \textit{Narrative}, \textit{Descriptive}, and \textit{Expository} writing. To ensure alignment with real-world educational scenarios, we collect and manually refine a total of 728 essay prompts. These prompts are further categorized into two types based on their instruction style: \textit{Open-Ended} and \textit{Constrained}, allowing us to examine LLMs' behavior under different writing conditions, as introduced in \textbf{Section}~\ref{Datasets}.
%Furthermore, to precisely assess how LLM performs in Chinese essay writing, we categorize these prompts into two sets according to their instruction requirements: \textit{Open-Ended} and \textit{Constrained}. 
Additionally, to overcome the limitations of existing evaluation methods for Chinese essay writing, we propose a fine-grained and genre-oriented evaluation framework, as shown in \autoref{fig:comparison}. We define multiple evaluation traits with hierarchical dependencies based on their complexity, ranging from basic to advanced requirements for each essay genre. For each trait, we design targeted sub-questions that reflect genre-specific writing expectations at different levels. To account for the hierarchical nature of these traits, we further introduce a dependency-weighted score aggregation strategy to better capture the writing quality, as introduced in \textbf{Section}~\ref{Protocol}.

We conduct two key experiments to validate the proposed framework. First, to assess its effectiveness and robustness, we perform a comprehensive human agreement study and a quality sensitivity analysis. The results demonstrate that our evaluation protocol aligns closely with human judgments, especially when applied to more advanced LLMs. More importantly, it significantly improves the ability to distinguish essay quality across high-, medium-, and low-level responses (See \textbf{Section}~\ref{Human-Agree}). Second, we benchmark 15 large-scale LLMs on the Chinese essay writing using our framework, offering detailed comparisons of their capabilities in writing Chinese essays (See \textbf{Section}~\ref{Benchmarking}).

In \autoref{tab:benchmark_comparison}, we highlight the key differences between our work and existing approaches. In summary, our main contributions are as follows:
\begin{itemize}
    \item We present \benchName, a multi-genre benchmark tailored for Chinese essay writing, covering \textit{Argumentative}, \textit{Narrative}, \textit{Descriptive}, and \textit{Expository} genres. The benchmark is curated from real-world scenarios and is suitable for practical use in educational applications.
    \item We propose an effective and robust evaluation protocol for Chinese essays that aligns closely with human judgments and greatly improves the ability to distinguish essays of varying quality.
    \item We benchmark 15 widely used large-scale LLMs to evaluate their strengths and weaknesses in Chinese essay writing, and highlight areas for future improvement.
\end{itemize}

\begin{figure*}
    \centering
    \includegraphics[width=\textwidth]{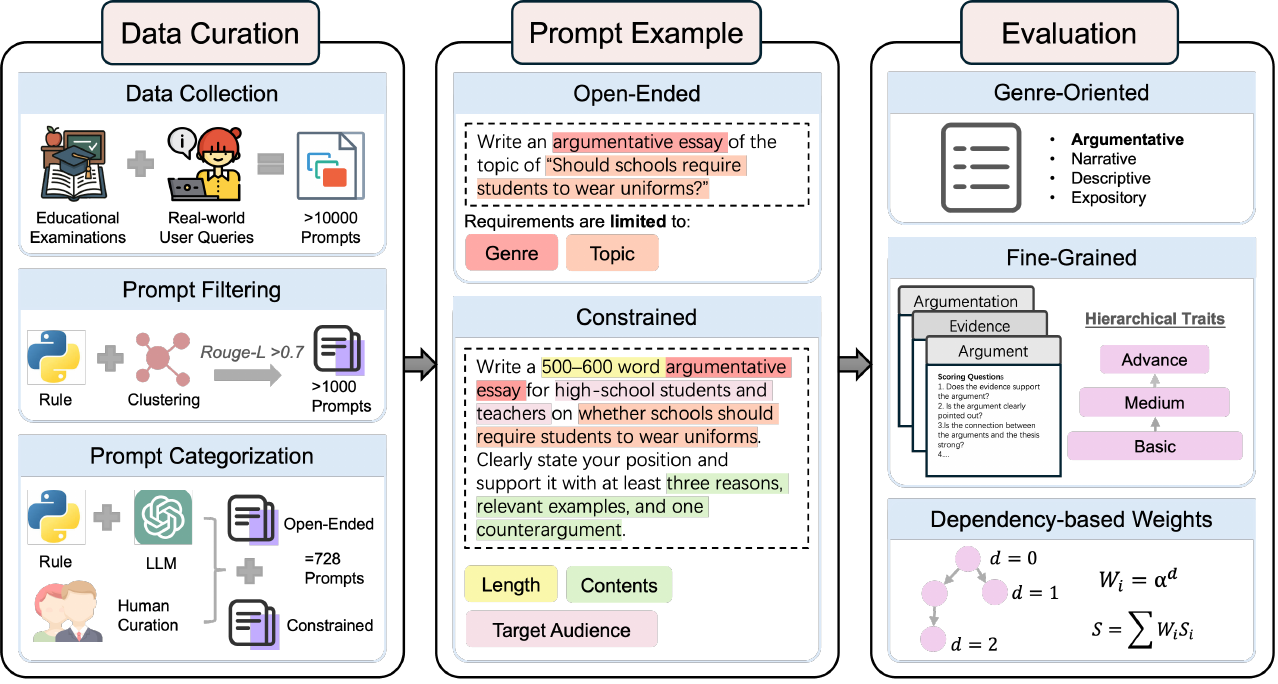}
    \caption{Overview of the \benchName dataset curation, representative prompts, and the evaluation framework.}
    \label{fig:overview}
\end{figure*}

\section{\benchName Dataset}\label{Datasets}

\benchName originally contributes to developing the datasets specifically tailored for Chinese Essay Writing. While prior benchmarks~\cite{wu2025writingbench} have largely provided queries on creative writing tasks in general domains, they do not adequately capture the structure, purpose, and constraints of formal essays, particularly within educational and academic contexts. To effectively benchmark the essay generation abilities, \benchName introduces a comprehensive set of essay prompts that span four major and widely recognized genres in Chinese writing instruction~\cite{chadbourne1983puzzling}: \textit{Argumentative}, \textit{Narrative}, \textit{Descriptive}, and \textit{Expository} essays, which cover the majority of Chinese prose compositions in educational settings. Furthermore, to support comprehensive evaluation, we categorize prompts into two distinct sets based on their multiple constraints. In this section, we describe the essay prompt construction process in detail, including data collection and quality control, and the two-phase query categorization procedures.

\subsection{Prompt Collection}

As shown in \autoref{fig:overview}, to reflect real-world usage and align with educational settings, we choose to collect prompts from practical and authentic resources. Specifically, we collect data from two primary resources, namely 1) real-world user queries obtained through online chatbot interactions, reflecting informal and user-generated prompts in tutoring or self-study contexts.
2) educational examination materials, including official Chinese essay prompts, represent standardized and curriculum-aligned writing tasks used in formal assessments.
%1) online chatbot interaction logs, which reflect informal and user-generated prompts in tutoring or self-study contexts, and 2) official Chinese essay examination materials, which represent standardized and curriculum-aligned writing tasks in formal assessments. 

\subsection{Prompt Filtering}
% We begin by collecting several thousand raw prompts from these sources to construct a broad candidate pool. 
Building on the collected prompts from these two sources, we construct a broad candidate pool containing several thousand raw entries.
To ensure the quality and representativeness of the datasets, we implement a multi-step filtering pipeline. First, we apply heuristic-based rules to remove irrelevant and low-quality prompts. We then employ clustering methods (e.g., $K$-means~\cite{hastie2009elements} with elbow method) to detect and eliminate duplicate or near-duplicate entries. To further enhance prompt diversity, we compute pairwise ROUGE-L scores between prompts and retain only those pairs with a similarity score below 0.7~\cite{jiang2024followbench}. In this stage, we get over 1000 relative prompts covering essay writing.

% In total, we get 728 prompts that capture a wide range of topics, genres, and instructional objectives in real-world Chinese writing tasks.

% We choose to collect data in real-world scenarios considering the educational settings. Generally, we collect prompt from two primary sources: online chatbot interaction logs and authentic Chinese essay examination materials, which represent distinct yet complementary perspectives on the essay writing tasks. Initially, we gather several thousand raw prompts from these sources to build a broad candidate pool. To enhance the quality and representativeness of the dataset, we apply a two-step filtering pipeline. This includes heuristic-based rules to remove the irrelevant and noisy prompt. Then we use the cluster methods to identify and eliminate the duplicate or near-duplicate prompts. This results in a refined and diverse set of prompts that better capture the range of topics, genres, and instructional goals in real-world Chinese tasks.

\subsection{Prompt Categorization}

%\paragraph{Prompt Composition.} 
To better evaluate how LLMs perform at different levels of writing difficulty, we divide the prompts into two subsets: \textit{Open-Ended} and \textit{Constrained}. %To distinguish between these two levels, 
To support this categorization, we first analyze the collected prompts and define five key factors that influence writing complexity and reader expectations: (1) Genre Specification. Each prompt clearly defines the target genres, including argumentative, narrative, descriptive, or expository, which guide the structural and rhetorical style of the expected response. (2) Topic Specification. Prompts indicate a central topic that the essay should focus on. For example, an argumentative prompt may require elaborating on a specific viewpoint, while an expository prompt asks for the introduction of a particular object or concept. (3) Content Constraints. These constraints specify required elements or themes within the essay. For instance, an argumentative prompt may instruct to include a historical example. (4) Length Requirements. Some prompts include explicit word or paragraph limits, adding structural constraints that impact the planning and execution of essay writing. (5) Target Audience. Prompts may specify the intended readership, such as middle school students or readers of a children's literary magazine, influencing the tone, vocabulary, and complexity of the writing. In particular, each prompt explicitly specifies both the writing genre and the topic, ensuring clarity in the contents.

\begin{figure}
    \centering
    \includegraphics[width=\linewidth]{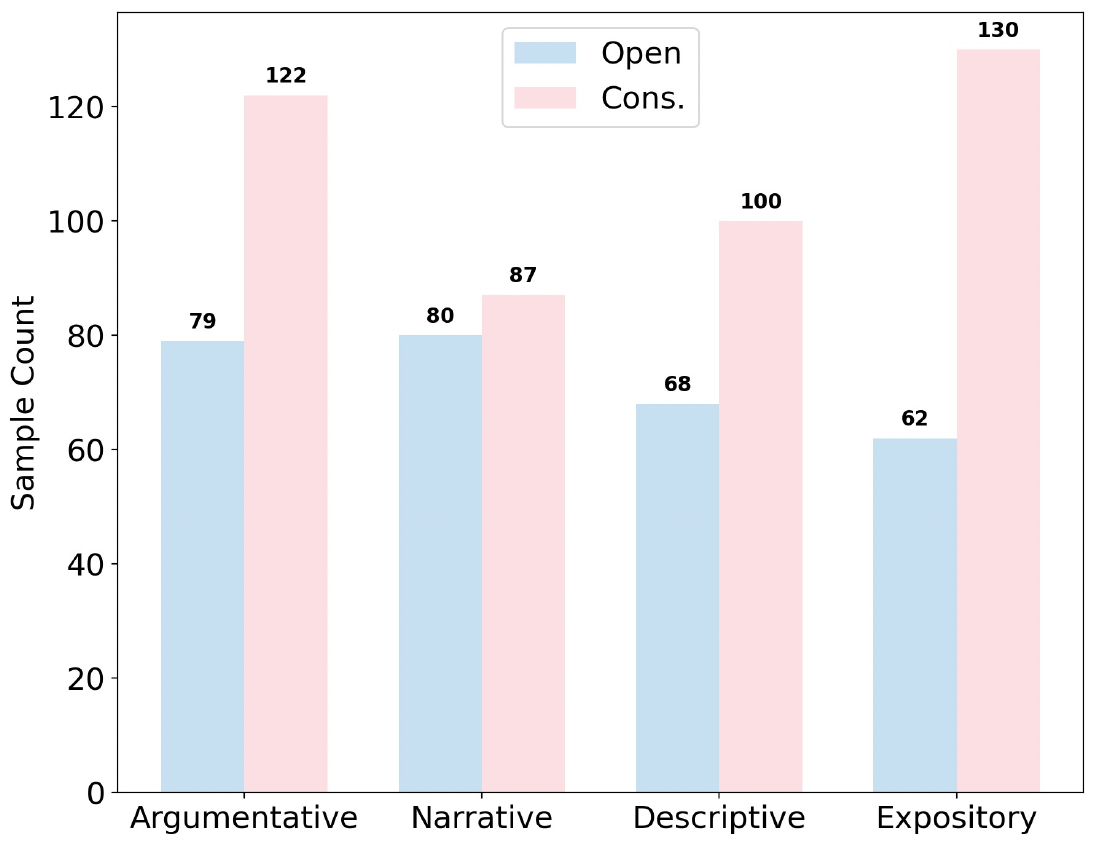}
    \caption{Dataset Statistics. Note that \textbf{Open} denotes Open-Ended sets, \textbf{Cons.} refers to Constrained sets.}
    \label{fig:stats}
\end{figure}

%\paragraph{Constraints Categorization.} 
%We categorize each prompt into either the \textit{Open-Ended} or \textit{Constrained} set based on its constraint composition. 
Building on the above-mentioned factors, we categorize each prompt into the either set based on the presence of constraints beyond the genre and topic, i.e., prompts in the \textit{Open-Ended} set include only basic instructions (genre and topic), while those in the \textit{Constrained} set contain additional requirements, such as length, content focus, or stylistic constraints. To perform this classification, we adopt a hybrid approach that combines rule-based parsing with LLM-based analysis. Specifically, rule-based methods are applied to identify explicit length constraints, while LLMs are used to detect more nuanced elements, such as topic- and content-related restrictions. All prompts are then manually reviewed by the authors to correct any misclassifications and ensure the overall consistency and quality of the dataset. After manual curation, we totally get 728 prompts that capture a wide range of topics, genres, and instructional objectives in real-world Chinese writing tasks. The statistics of the dataset are shown in \autoref{fig:stats}.

 % To begin, we assess the difficulty of each factor individually, classifying them into either easy or hard level based on predefined rules. Notably, we exclude Genre from this evaluation, as genre type alone does not directly indicate task difficulty in our framework.

 % Each factor is then assigned a binary difficulty score (0 for easy, 1 for hard), and the scores are summed up for each prompt. If the total score is 0 or 1, the prompt is categorized as \textit{Easy}; if the score is 2 or higher, it is classified as \textit{Hard}. Such methods provide a systematic and interpretable way to assess and categorize the prompt difficulty.

\section{\benchName Evaluation Protocol}\label{Protocol}

In this section, we present the design of our evaluation framework for assessing Chinese essays. Due to the open-ended and reference-free nature of essay writing, we adopt the LLM-as-a-judge paradigm~\cite{chen-etal-2024-humans,gu2025surveyllmasajudge} as our evaluation approach. Despite its growing popularity, existing protocols for evaluating essay generation remain insufficient, particularly in the context of Chinese writing, which involves distinct linguistic features and culturally rooted rhetorical conventions~\cite{liu-etal-2024-cerd}. To meet these evaluation needs, we propose a genre-oriented, fine-grained, and dependency-aware evaluation framework for \benchName, capturing structural, linguistic, and hierarchical aspects of Chinese essays. 

%tailored to the unique characteristics of Chinese essays. 

\noindent \textbf{Genre-Oriented Evaluation.} In practical essay evaluation, the criteria for assessing quality often vary across genres, as different genres emphasize distinct aspects of writing based on their inherent characteristics. As a result, our framework is adapted to different genres accordingly. Following the principal rubrics outlined in~\cite{blanchard2013toefl11, asap-aes}, we refine and construct genre-specific evaluation traits that align with Chinese writing conventions. Specifically, we define six genre-specific evaluation dimensions, each designed with expectations that range from basic to advanced requirements, as detailed in Appendix~\ref{appendix:traits}. This setup allows our framework to effectively capture the distinctive features of different essay types and evaluate essays across varying quality levels. 

% For argumentative essays, logical structure and reasoning are key; for narrative essays, the coherence of the storyline and the richness of character descriptions are emphasized; for descriptive essays, the vividness and sensory imagery are crucial; and for expository essays, clarity, factual accuracy, and coherence of explanation are prioritized.

\begin{table*}[t]
\centering
\small
\setlength{\tabcolsep}{9pt}
\begin{tabular}{l|cc|cc|cc|cc|cc}
\toprule
\multirow{2}{*}{Methods} & \multicolumn{2}{c}{Overall}  & \multicolumn{2}{c}{Argumentative} & \multicolumn{2}{c}{Narrative} & \multicolumn{2}{c}{Descriptive} & \multicolumn{2}{c}{Expository} \\
\ & $\rho$ & $\tau$ \ & $\rho$ & $\tau$ & $\rho$ & $\tau$ & $\tau$ & $\rho$ & $\rho$ & $\tau$ \\
\midrule
\ & \multicolumn{10}{c}{\cellcolor{gray!20}\textcolor{blue}{\textit{\textbf{DeepSeek-V3}}}} \\
% \cmidrule(lr){6-7}
Align-Score & \textbf{0.674} & \textbf{0.599} & \textbf{0.744} & 
\textbf{0.674} & 0.635 & \textbf{0.559} & 0.656 & 0.580 & \textbf{0.656} & \textbf{0.578}\\

Ours w/o \textit{WT}. & 0.646 & 0.529 & 0.701 & 0.576 & 0.596 &0.464 & 0.778 & 0.672 & 0.509 & 0.405 \\

Ours & 0.667 & 0.549 & 0.670 & 0.546 & \textbf{0.648} & 0.518 & \textbf{0.796} & \textbf{0.676} & 0.554 & 0.458\\

\midrule
\ & \multicolumn{10}{c}{\cellcolor{gray!20}\textcolor{blue}{\textit{\textbf{GPT-4o}}}} \\
Align-Score & 0.628  & 0.546  & 0.587 & 0.516 & 0.582 & 0.514 & 0.642 & 0.563 & 0.700 & 0.594\\
Ours w/o \textit{WT}. & 0.706 & 0.596 &  0.747 & 0.643 & 0.747 & 0.645 & 0.688 & 0.576 & 0.643 & 0.520\\
Ours & \textbf{0.733} & \textbf{0.627} & \textbf{0.754} & \textbf{0.662} & \textbf{0.773} & \textbf{0.658} & \textbf{0.700} & \textbf{0.594} & \textbf{0.707} & \textbf{0.601}\\
\midrule
\ & \multicolumn{10}{c}{\cellcolor{gray!20}\textcolor{blue}{\textit{\textbf{DeepSeek-R1}}}} \\
Align-Score & 0.749 & 0.667 & 0.745 & 0.667 &0.764 & 0.695  & 0.709 & 0.617 & 0.778  & 0.686\\
Ours w/o \textit{WT}. & 0.803 & 0.685 & 0.789 & 0.648 & 0.830 & 0.719 & 0.817 & 0.702 & 0.785 & 0.669 \\
Ours & \textbf{0.816} & \textbf{0.704} & \textbf{0.795} & \textbf{0.673} & \textbf{0.838} & \textbf{0.724} & \textbf{0.839} & \textbf{0.731} & \textbf{0.791} & \textbf{0.690}  \\

\bottomrule
\end{tabular}
\caption{Comparison of human agreement evaluation across different scoring methods on sampled data. $\rho$ refers to Spearman's $\rho$, $\tau$ denotes the Kendall's $\tau$, while \textit{WT.} represents the dependency-based weights.}
\label{tab:aggreement_score}
\end{table*}

\noindent \textbf{Fine-Grained Evaluation.} Existing methods to evaluate individual dimensions typically rely on direct scoring or binary questions~\cite{que2024hellobench}, but these approaches are often limited by their coarse granularity~\cite{kim-etal-2025-biggen}. Inspired by the multi-trait evaluation design~\cite{lee2024unleashing}, we introduce a set of sub-questions ($q_i$) under each evaluation dimension to enable more nuanced assessments (See Appendix~\ref{appendix: prompts}). We adopt the Chain-of-Thought (CoT)~\cite{wei2023chainofthoughtpromptingelicitsreasoning} prompting technique to guide LLMs in analyzing responses and identifying linguistic evidence in support of the assigned scores. The final score for $t$-th dimension $S_{t}$ is computed by aggregating the scores of the corresponding sub-questions as $S_t = \sum\limits_{i}{q_i}$. 
% \begin{equation}
% \small
% S_t = \sum\limits_{i}{q_i}
% \end{equation}

\noindent \textbf{Dependency-Aware Evaluation.} Many existing works determine the overall response quality by simply summing or averaging the scores of individual dimensions. However, based on our observations and preliminary experiments, we find that hierarchical traits contribute unequally, and treating them independently often fails to capture nuanced features in high-quality essays. To address this limitation, we propose a dependency-aware scoring approach inspired by~\cite{saaty1980analytic, zizovic2019new}, which assigns weights to each trait based on its position in the evaluation hierarchy. For example, traits at the base level are assigned a depth ($d$) of $0$, while mid-level traits have a depth of $1$. The weights ($W_t$) are computed using \autoref{eq:weight}, with the hyperparameter $\alpha$ controlling the importance of basic and advanced levels. The final score is a weighted sum of all trait scores.
\begin{equation}
    W_{t} = \alpha^{d}.
\label{eq:weight}
\end{equation}

\begin{table*}[t]
\centering
\small
\setlength{\tabcolsep}{11pt}
\begin{tabular}{l|cc|cc|cc}
\toprule
 \multirow{2}{*}{\textbf{Method}} & \multicolumn{2}{c}{\textbf{DeepSeek-V3}} & \multicolumn{2}{c}{\textbf{GPT-4o}} & \multicolumn{2}{c}{\textbf{DeepSeek-R1}} \\
 \ & $U_{p}\uparrow$ & $MD_{std}\uparrow$  & $U_{p}\uparrow$ & $MD_{std}\uparrow$ & $U_{p}\uparrow$ & $MD_{std}\uparrow$ \\
\midrule
\ & \multicolumn{6}{c}{\cellcolor{gray!20}\textcolor{blue}{\textit{\textbf{high\&medium}}}} \\
% \hline
Align-Score & $0.56_{<0.05}$ & $0.17_{0.62}$ & $0.56_{=0.14}$  & $0.25_{0.84}$ & $0.64_{=1.43}$ & $0.42_{0.77}$\\
Ours & $\bm{0.57}_{<0.10}$ &  $\bm{0.24}_{0.74}$ & $\bm{0.66}_{<0.05}$ & $\bm{0.45}_{1.05}$ & $\bm{0.79}_{<0.05}$ & $\bm{0.70}_{0.79}$\\
\midrule
\ & \multicolumn{6}{c}{\cellcolor{gray!20}\textcolor{blue}{\textit{\textbf{medium\&low}}}}\\
% \hline
Align-Score & $\bm{0.90}_{<0.05}$ &  $1.42_{1.26}$& $0.87_{<0.05}$ & $2.16_{1.48}$ & $0.93_{<0.05}$  & $1.98_{1.05}$ \\
Ours   & $0.78_{<0.05}$ &  $\bm{1.96}_{1.41}$   & $\bm{0.93}_{<0.05}$  & $\bm{2.46}_{1.42}$ & $\bm{0.97}_{<0.05}$ & $\bm{2.79}_{1.32}$\\
\midrule
\ & \multicolumn{6}{c}{\cellcolor{gray!20}\textcolor{blue}{\textit{\textbf{high\&low}}}} \\
% \hline
Align-Score  & $\bm{0.92}_{<0.05}$ & $1.66_{1.41}$ &  $0.93_{<0.05}$ & $2.41_{1.35}$  & $0.97_{<0.05}$ & $2.41_{1.06}$\\

Ours & $0.82_{<0.05}$  & $\bm{2.13}_{1.42}$  & $\bm{0.98}_{<0.05}$ & $\bm{2.90}_{1.41}$ &  $ \bm{0.99}_{<0.05}$ & $\bm{3.49}_{1.37}$\\
\bottomrule
\end{tabular}
\caption{Comparison of sensitivity analysis results between baselines and our proposed evaluation method, with the best-performing scores highlighted in bold. $p$ denotes statistical significance, and $std$ indicates standard deviation.}
\label{tab:sensitivity_score}
\end{table*}

\section{Human Agreement Evaluation}\label{Human-Agree}

To validate the effectiveness of our evaluation protocol, we conduct a comprehensive human agreement study in Chinese essays. Specifically, the study focuses on two aspects: \textbf{1) Ranking Agreement}, which measures how closely the rankings produced by our evaluation framework align with human judgments; and \textbf{2) Sensitivity Evaluation}, which assesses the robustness of the framework in distinguishing essays of varying quality.

\subsection{Experiment Setup}

\noindent \textbf{Datasets.} We randomly sample 80 prompts across different categories, selecting ten prompts per genre per difficulty level. For each prompt, we evaluate essays generated by seven language models, including both open- and closed-source models: LLaMA-3.1-70B-Instruct~\cite{grattafiori2024llama3herdmodels}, Qwen-2.5-72B-Instruct~\cite{qwen2025qwen25technicalreport}, GPT-3.5-turbo~\cite{brown2020language}, Claude-3.5-Sonnet~\cite{ouyang2022traininglanguagemodelsfollow}, Deepseek-v3~\cite{deepseekai2025deepseekv3technicalreport}, Grok-3~\cite{xai2025grok3}, and GPT-4o~\cite{openai2024gpt4ocard}. We then recruit 14 professional annotators with rich backgrounds in Chinese linguistics to assess the generated essays. To ensure reliability and consistency, we adopt a pairwise comparison annotation method~\cite{wen2024benchmarking}, assigning each essay pair to three annotators. In total, the annotation process results in 5,040 labeled data. Finally, in \autoref{tab:kappa_score}, a Fleiss' Kappa Agreement~\cite{fleiss1971measuring} is used to measure the agreements among three evaluators to ensure the annotation quality.

\begin{table}[t]
\centering
\small
\setlength{\tabcolsep}{2pt}
\begin{tabular}{c|c|c|c|c}
\toprule
\multicolumn{5}{c}{\cellcolor{gray!20}\textcolor{blue}{\textit{\textbf{Pair-Wise Kappa Score}}}} \\
\midrule

  Overall&  Argumentative & Narrative & Descriptive & Expository \\
\midrule
0.469 & 0.477 & 0.457 & 0.464 & 0.475 \\
\bottomrule
\end{tabular}
\caption{Fleiss' Kappa Agreement on pairwise annotations. A score between 0.41 to 0.60 indicates moderate inter-annotator agreement~\cite{qin-etal-2024-infobench}.}
\label{tab:kappa_score}
\end{table}

\noindent\textbf{Baselines.} As the first to propose an evaluation protocol specifically tailored for Chinese essay writing, we compare our method against two baseline approaches: (1) \textbf{Align Scoring}~\cite{liu-etal-2024-alignbench} from AlignBench, which evaluates general Chinese writing quality, particularly, we slightly modify it to evaluate reference-free essays; and (2) \textbf{Ours w/o Weights}, which applies the same evaluation rubrics as our method but without dependency-based weighting.

\noindent \textbf{Judges.} To verify how well the proposed evaluation method works, we employ three LLMs as judges, including DeepSeek-V3~\cite{deepseekai2025deepseekv3technicalreport}, DeepSeek-R1~\cite{deepseekai2025deepseekr1incentivizingreasoningcapability} and GPT-4o~\cite{openai2024gpt4ocard} to assign scores 1$\sim$10 to each sub-question within every evaluation trait. Each model analyzes all sub-questions in a single turn. Specifically, we convert annotated pairwise comparisons into model rankings using a voting-based scoring approach to facilitate more effective comparisons. In all experiments, the temperature is set to $0.2$, and the parameter $\alpha$ is fixed to $3$.
% This comparison allows us to assess our hierarchical weighting strategy over more conventional scoring methods.
\subsection{Ranking Agreement}
To assess the ranking agreement, we use \textbf{Spearman's Rank Correlation}~\cite{spearman1904association} and \textbf{Kendall's $\tau$}~\cite{kendall1938rank}, which capture monotonic relationships between rankings. As shown in \autoref{tab:aggreement_score}, our fine-grained and genre-oriented evaluation framework shows strong alignment with human judgments~\cite{shen-etal-2023-large}, achieving high correlations in both Spearman's $\rho$ and Kendall's $\tau$. 
From these results, we draw three key conclusions: \textbf{(1) Our protocol performs better with stronger LLMs.} Our method crafts dimension-specific sub-questions and uses the CoT strategy to analyze the text and then assign all scores in a single turn. More powerful models exhibit a superior understanding of this complex and fine-grained process. Notably, DeepSeek-R1 achieves an almost perfect alignment with human annotations, with $\rho=0.816$ and $\tau=0.704$. 
\textbf{(2) Dependency-based score aggregation improves performance by approximately $2\%$.} Incorporating trait-level weights consistently improves alignment across different judges and essay genres, indicating that when assessing essays, the higher-level dimensions contribute more significantly to accurate evaluation. \textbf{(3) Our framework achieves higher alignment in Narrative and Descriptive genres.} Unlike argumentative and expository essays that emphasize logical structure and coherence and are effectively handled by general text evaluation method, narrative and descriptive writing focus on vivid imagery, rhetorical richness, and lexical complexity, which benefit more from our evaluation approach.

% Claude-3.5-sonnet-2410~\cite{anthropic2024claude35sonnet}, Claude-3-7-sonnet-25-2~\cite{anthropic2025claude37sonnet}, Grok-2~\cite{xai2024grok2}, Grok-3~\cite{xai2025grok3},  GPT-3.5-turbo~\cite{brown2020language}, GPT-4o-latest and GPT-4o-mini~\cite{openai2024gpt4ocard},, Gemini-2.0-flash~\cite{geminiteam2025geminifamilyhighlycapable}, and LLaMa-3.1-70B-Instruct and LLaMA-3.3-70B-Instruct~\cite{grattafiori2024llama3herdmodels}; as well as Chinese Language Models: Qwen-max-25-01~\cite{qwen2025qwen25max}, Qwen-2.5-72B-Instruct~\cite{qwen2025qwen25technicalreport},
% ChatGLM-turbo~\cite{glm2024chatglmfamilylargelanguage}, DeepSeek-V3~\cite{deepseekai2025deepseekv3technicalreport}, and Doubao-1.5-pro~\cite{doubao2025}. 

\begin{table*}
\centering
\small
\setlength{\tabcolsep}{6pt}
\begin{tabular}{l|c|cc|cc|cc|cc}
\toprule
\multirow{2}{*}{\textbf{Models}} & \multirow{2}{*}{\textbf{Overall}}  & \multicolumn{2}{c}{\textbf{Argumentative}} & \multicolumn{2}{c}{\textbf{Narrative}} & \multicolumn{2}{c}{\textbf{Descriptive}} & \multicolumn{2}{c}{\textbf{Expository}} \\
\ & \ & Open & Cons. & Open & Cons. & Open & Cons. & Open & Cons. \\
\midrule
\ & \multicolumn{9}{c}{\textit{English Language Models}} \\
\midrule
\cellcolor{Tan!10}Claude-3.7-sonnet~\cite{anthropic2025claude37sonnet}  & \cellcolor{Tan!10}\textbf{76.6} & \cellcolor{Tan!10}\textbf{77.7} & \cellcolor{Tan!10}\textbf{78.8} & \cellcolor{Tan!10}\textbf{75.7} & \cellcolor{Tan!10}\textbf{75.3} & \cellcolor{Tan!10}\underline{74.6} & \cellcolor{Tan!10}\textbf{73.6} &\cellcolor{Tan!10}\underline{77.5} & \cellcolor{Tan!10}79.0 \\

\cellcolor{Tan!10}Claud-3.5-sonnet~\cite{anthropic2024claude35sonnet} & \cellcolor{Tan!10}75.4 & \cellcolor{Tan!10}73.4 & \cellcolor{Tan!10}73.8 & \cellcolor{Tan!10}\underline{75.3} & \cellcolor{Tan!10}73.6 & \cellcolor{Tan!10}\textbf{74.8} & \cellcolor{Tan!10}73.4 & \cellcolor{Tan!10}77.1 & \cellcolor{Tan!10}\textbf{80.4} \\

\cellcolor{Tan!10}Grok-2~\cite{xai2024grok2} & \cellcolor{Tan!10}75.3 & \cellcolor{Tan!10}75.6 & \cellcolor{Tan!10}78.5 & \cellcolor{Tan!10}71.5 & \cellcolor{Tan!10}73.6 & \cellcolor{Tan!10}70.2 & \cellcolor{Tan!10}\underline{73.5} & \cellcolor{Tan!10}75.1 & \cellcolor{Tan!10}79.3 \\

\cellcolor{Tan!10}Grok-3~\cite{xai2025grok3} & \cellcolor{Tan!10}74.6 & \cellcolor{Tan!10}74.9 & \cellcolor{Tan!10}78.1 & \cellcolor{Tan!10}73.6 & \cellcolor{Tan!10}72.8 & \cellcolor{Tan!10}73.1 & \cellcolor{Tan!10}72.0 & \cellcolor{Tan!10}73.3 & \cellcolor{Tan!10}76.4 \\

\cellcolor{Tan!10}GPT-4o~\cite{openai2024gpt4ocard} & \cellcolor{Tan!10}74.2 & \cellcolor{Tan!10}74.8 & \cellcolor{Tan!10}76.9 & \cellcolor{Tan!10}72.8 & \cellcolor{Tan!10}72.4 & \cellcolor{Tan!10}70.5 & \cellcolor{Tan!10}71.7 & \cellcolor{Tan!10}75.8 & \cellcolor{Tan!10}76.7 \\

\cellcolor{Tan!10}GPT-4o-mini~\cite{openai2024gpt4ocard} & \cellcolor{Tan!10}71.7 & \cellcolor{Tan!10}72.0 & \cellcolor{Tan!10}74.1 & \cellcolor{Tan!10}71.6 & \cellcolor{Tan!10}68.4 & \cellcolor{Tan!10}69.9 & \cellcolor{Tan!10}65.9 & \cellcolor{Tan!10}72.8 & \cellcolor{Tan!10}76.7 \\

\cellcolor{Tan!10}GPT-3.5-turbo~\cite{brown2020language} &\cellcolor{Tan!10}51.5 & \cellcolor{Tan!10}49.4 &  \cellcolor{Tan!10}51.4 & \cellcolor{Tan!10}56.5 & \cellcolor{Tan!10}53.1 & \cellcolor{Tan!10}51.1 & \cellcolor{Tan!10}46.8 & \cellcolor{Tan!10}50.0 & \cellcolor{Tan!10}52.9 \\

\cellcolor{Tan!10}Gemini-2.0-flash~\cite{geminiteam2025geminifamilyhighlycapable} & \cellcolor{Tan!10}72.9\cellcolor{Tan!10} & \cellcolor{Tan!10}74.5 & \cellcolor{Tan!10}76.3 & \cellcolor{Tan!10}71.5 & \cellcolor{Tan!10}71.1 & \cellcolor{Tan!10}68.4 & \cellcolor{Tan!10}67.6 & \cellcolor{Tan!10}76.7 & \cellcolor{Tan!10}75.4 \\

\cellcolor{Tan!10}LLaMa-3.3-70B~\cite{grattafiori2024llama3herdmodels} & \cellcolor{Tan!10}61.4 & \cellcolor{Tan!10}61.2 & \cellcolor{Tan!10}64.1 & \cellcolor{Tan!10}62.3 & \cellcolor{Tan!10}60.3 & \cellcolor{Tan!10}56.2 & \cellcolor{Tan!10}53.8 & \cellcolor{Tan!10}63.2 &\cellcolor{Tan!10}67.1 \\

\cellcolor{Tan!10}LLaMa-3.1-70B~\cite{grattafiori2024llama3herdmodels} & \cellcolor{Tan!10}40.5 & \cellcolor{Tan!10}37.6 & \cellcolor{Tan!10}46.6 & \cellcolor{Tan!10}35.1 & \cellcolor{Tan!10}28.6 & \cellcolor{Tan!10}45.0 & \cellcolor{Tan!10}42.2 & \cellcolor{Tan!10}39.6 & \cellcolor{Tan!10}44.8 \\
\midrule
\ & \multicolumn{9}{c}{\textit{Chinese Language Models}} \\
\midrule
\cellcolor{BlueGreen!10}Qwen-Max~\cite{qwen2025qwen25max} 
   & \cellcolor{BlueGreen!10}\underline{75.6 }& \cellcolor{BlueGreen!10}74.5 & \cellcolor{BlueGreen!10}\underline{78.7} & \cellcolor{BlueGreen!10}73.5 & \cellcolor{BlueGreen!10}\underline{74.7} & \cellcolor{BlueGreen!10}74.1 & \cellcolor{BlueGreen!10}72.6 & \cellcolor{BlueGreen!10}77.1 & \cellcolor{BlueGreen!10}77.6\\

\cellcolor{BlueGreen!10}Qwen2.5-72B-Instruct~\cite{qwen2025qwen25technicalreport} & \cellcolor{BlueGreen!10}72.7 & \cellcolor{BlueGreen!10}73.1 & \cellcolor{BlueGreen!10}75.2 & \cellcolor{BlueGreen!10}71.7 & \cellcolor{BlueGreen!10}71.4 & \cellcolor{BlueGreen!10}68.8 & \cellcolor{BlueGreen!10}68.8 & \cellcolor{BlueGreen!10}74.5 & \cellcolor{BlueGreen!10}75.5\\

\cellcolor{BlueGreen!10}DeepSeek-V3~\cite{deepseekai2025deepseekv3technicalreport} & \cellcolor{BlueGreen!10}75.1 & \cellcolor{BlueGreen!10}\underline{77.2} & \cellcolor{BlueGreen!10}77.9  & \cellcolor{BlueGreen!10}71.2 & \cellcolor{BlueGreen!10}71.8 & \cellcolor{BlueGreen!10}72.7 & \cellcolor{BlueGreen!10}67.8 & \cellcolor{BlueGreen!10}\textbf{80.4} & \cellcolor{BlueGreen!10}\underline{79.4} \\

\cellcolor{BlueGreen!10}Doubao-1.5~\cite{doubao2025} & \cellcolor{BlueGreen!10}73.3 & \cellcolor{BlueGreen!10}75.1 & \cellcolor{BlueGreen!10}76.2   & \cellcolor{BlueGreen!10}72.4 & \cellcolor{BlueGreen!10}70.8 & \cellcolor{BlueGreen!10}70.8 & \cellcolor{BlueGreen!10}69.5 & \cellcolor{BlueGreen!10}75.4 & \cellcolor{BlueGreen!10}75.1 \\

\cellcolor{BlueGreen!10}ChatGLM-turbo~\cite{glm2024chatglmfamilylargelanguage} & \cellcolor{BlueGreen!10}71.2 & \cellcolor{BlueGreen!10}70.0 & \cellcolor{BlueGreen!10}70.8 & \cellcolor{BlueGreen!10}70.0 & \cellcolor{BlueGreen!10}69.6 & \cellcolor{BlueGreen!10}69.2 & \cellcolor{BlueGreen!10}68.7 & \cellcolor{BlueGreen!10}74.2 & \cellcolor{BlueGreen!10}75.8 \\

\bottomrule
\end{tabular}
\caption{Benchmarking Results on Chinese Essay Writing. In each column, the highest and the second highest performance is highlighted in \textbf{bold} and is \underline{underlined}. \textbf{Open} denotes Open-Ended and \textbf{Cons.} denotes Constrained.}
\label{tab:benchmarking}
\end{table*}

\subsection{Sensitivity Analysis}

Accurately determining an LLM's proficiency in specific capabilities is essential for identifying its limitations and guiding improvements~\cite{kim-etal-2025-biggen}. Therefore, it is crucial that the evaluation protocol reliably reflects both high- and low-quality output. To this end, we conduct a sensitivity analysis to examine how effectively our evaluation protocol distinguishes essays of varying quality.

Accordingly, we categorize the essays into three quality tiers: high-, medium-, and low-quality based on the top-ranked, median-ranked, and bottom-ranked essays from the manually annotated data. Then we apply \textbf{Mann-Whitney $U$ test}~\cite{mann1947test} and compute the \textbf{Mean Difference} ($MD$) to assess the robustness of the methods, as shown in \autoref{tab:sensitivity_score}. Take the \textit{high\&medium} set as an example. The $U$ score indicates the proportion of cases in which high-quality data receive a higher score than medium-quality data. The mean difference reflects the average score difference between the high- and medium-quality data.

The sensitivity analysis in \autoref{tab:sensitivity_score} shows that \textbf{our evaluation method is effective at distinguishing essays of varying quality compared to the baseline}. Notably, our method shows significantly better performance in the high- and medium-quality essay classification, with an improvement ranging from approximately $2\%$ to $10\%$. Furthermore, it yields a larger mean difference, suggesting that the score distributions between quality levels are more distinguishable. These trends hold consistently across all judge models, highlighting the robustness and sensitivity of our framework when evaluating outputs from strong LLMs. Overall, R1 emerges as the top-performing model, achieving the highest $U$ score and exhibiting a pronounced distinction across all quality levels.

% \begin{figure*}
%     \centering
%     \includegraphics[width=0.6\linewidth]{figures/performance.pdf
%     }
%     \caption{Comparison of LLMs Performance across (a) different genres and (b) different sets.}
%     \label{fig:performance}
% \end{figure*}

\section{Benchmarking}\label{Benchmarking}

% Claude-3.5-sonnet-2410~\cite{anthropic2024claude35sonnet}, Claude-3-7-sonnet-25-2~\cite{anthropic2025claude37sonnet}, Grok-2~\cite{xai2024grok2}, Grok-3~\cite{xai2025grok3},  GPT-3.5-turbo~\cite{brown2020language}, GPT-4o-latest and GPT-4o-mini~\cite{openai2024gpt4ocard},, Gemini-2.0-flash~\cite{geminiteam2025geminifamilyhighlycapable}, and LLaMa-3.1-70B-Instruct and LLaMA-3.3-70B-Instruct~\cite{grattafiori2024llama3herdmodels}; as well as Chinese Language Models: Qwen-max-25-01~\cite{qwen2025qwen25max}, Qwen-2.5-72B-Instruct~\cite{qwen2025qwen25technicalreport},
% ChatGLM-turbo~\cite{glm2024chatglmfamilylargelanguage}, DeepSeek-V3~\cite{deepseekai2025deepseekv3technicalreport}, and Doubao-1.5-pro~\cite{doubao2025}. 

\subsection{Experiment Setup}

\paragraph{Baselines.} To explore how current state-of-the-art LLMs perform in Chinese essay writing, we meticulously select 15 popular large-scale LLMs for evaluation, including English language models and Chinese language models. We access proprietary LLMs via their official APIs and open-source LLMs through their public repositories. During writing, we set the temperature to 0.8 to encourage creativity in generation. 

\paragraph{Metrics.} Considering the inference time cost and overall performance, we adopt GPT-4o as the evaluation judge model. The temperature is set to 0.2 to ensure deterministic output, while all other parameters remain in their default settings. To facilitate fair comparison across models, we normalize the aggregated scores to a 100-point scale.

\paragraph{Main Results.} The benchmark results are presented in \autoref{tab:benchmarking}. Notably, state-of-the-art proprietary models achieve strong performance on the Chinese essay writing task, with Claude-3.7-sonnet attaining the highest overall score. Moreover, most newer versions outperform their predecessors, with the exception of Grok, as Grok-3 places greater emphasis on reasoning. It is worth highlighting that Chinese LLM families also perform competitively: Qwen-max ranks as the second-best model, DeepSeek surpasses Grok-3 and GPT-4o on this task, and Qwen-2.5-72B-Instruct outperforms both the GPT-4o-mini and its similarly sized counterpart, LLaMA-3.1-70B-Instruct.

\paragraph{Genre-based Performance.} \textbf{LLMs demonstrate stronger capabilities in writing argumentative and expository essays, while they fall short in narrative and descriptive genres} (see Fig~\ref{fig:genre_performance}). This disparity likely stems from the inherent characteristics of these genres: argumentative and expository essays emphasize structural coherence, logical reasoning, and clear topic development, where LLMs typically excel. In contrast, narrative and descriptive essays require creativity, emotional nuance, and context-aware storytelling. These challenges are further amplified in Chinese writing, where expressive richness, metaphorical language, and cultural context play a more significant role, especially in narrative and descriptive forms. Such features are difficult to model with LLMs, leading to degraded performance in these genres.

\begin{figure}
    \centering
    \includegraphics[width=\linewidth]{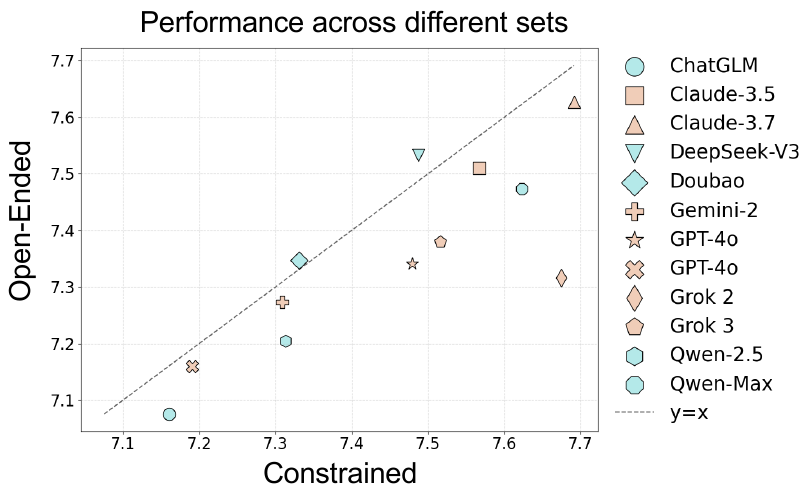}
    \caption{Comparison of Performance by Sets.}
    \label{fig:scatter}
\end{figure}

\paragraph{Open-Ended versus Constrained.} Interestingly, LLMs perform better in constrained sets than open-ended sets, as shown in \autoref{fig:scatter}. This is likely because constrained prompts provide more explicit requirements and clearer guidance, which help the models organize content, maintain relevance, and follow a well-defined structure. In contrast, open-ended prompts offer greater flexibility but less direction, placing higher demands on the model's ability to plan, generate diverse content, and maintain coherence without external constraints.

% \begin{figure*}
%     \centering
%     \includegraphics[width=0.8\linewidth]{figures/performance.pdf
%     }
%     \caption{Comparison of LLMs Performance across (a) different genres and (b) different sets.}
%     \label{fig:performance}
% \end{figure*}

% \begin{table*}[t]
% \small
% \centering
% \renewcommand{\arraystretch}{1.2}
% \setlength{\tabcolsep}{3.6pt}  % 调整列间距
% \begin{tabular}{lc|ccc|ccc}
% \toprule
% \multirow{2}{*}{\textbf{Benchmark}} & \multirow{2}{*}{\textbf{Num.}} & \multicolumn{3}{c|}{\textbf{Dataset Composition}} & \multicolumn{3}{c}{\textbf{Evaluation Method}} \\
%  \ & \ & Domain & Language & Constraints & LLM & \textit{Hier}. Traits & Aggregation \\
% \midrule
% C-Eval~\cite{huang2023ceval} & 13,948 & General & ZH & \xmark & \xmark & \xmark & - \\
% AlignBench~\cite{liu-etal-2024-alignbench} & 683 & General & ZH & \xmark & \cmark & \xmark & Direct \\
% LongBench-Write~\cite{bai2024longwriter} & 120 &  General Writing & ZH\&EN & \xmark & \cmark & \cmark & Direct \\
% HelloBench~\cite{que2024hellobench} & 647 & General & EN & \xmark & \cmark & \xmark & Weighted\\
% WritingBench~\cite{wu2025writingbench} & 1239 & General Writing & ZH\&EN & \cmark & \cmark & \xmark & Direct\\
% \midrule
% \textbf{\benchName (Ours)} & \textbf{728} & \textbf{Essay Writing} & \textbf{ZH} & \cmark & \cmark & \cmark & \textbf{Weighted} \\
% \bottomrule
% \end{tabular}
% \caption{Comparisons between \benchName and other benchmarks, illustrating the features including dataset size, dataset compositions, and evaluation methods.}
% \label{tab:benchmark_comparison}
% \end{table*}

\section{Related Work}

\noindent\textbf{LLM Generation Evaluation.} The rapid progress of LLMs prompts the need for a comprehensive evaluation of their text generation~\cite{liu-etal-2023-g, kim-etal-2025-biggen}. Existing frameworks are often task-specific: instruction-following is assessed via diverse prompts and constraint scenarios~\cite{qin-etal-2024-infobench, wen2024benchmarking, jiang2024followbench}, while reasoning is tested through multi-hop question answering~\cite{krishna2024fact, ling2025longreason}. In this work, we turn our attention to the issue of generated text quality evaluation. Previous research has addressed quality assessment in specific contexts: e.g., summarization~\cite{liu-etal-2024-benchmarking}, financial content~\cite{islam2023financebench, xie2024finben}, Wikipedia-style writing~\cite{gao2024evaluating, zhang2025wikigenbench}, and long-form text~\cite{tan2024proxyqa, que2024hellobench}. In contrast, we address the underexplored challenge of evaluating Chinese writing across literary genres, offering a systematic framework for multilingual LLM assessment.

\noindent \textbf{Automatic Essay Evaluation.} Automated Essay Scoring (AES) uses computer systems to assess written text in educational settings~\cite{dikli2006overview, attali2006automated}. While datasets like ASAP~\cite{asap-aes} and TOEFL11~\cite{blanchard2013toefl11} provide valuable English essay prompts, they are limited in scale and unsuitable for assessing LLM-generated essays, especially in non-English contexts. AES methods have progressed from hand-crafted features~\cite{yannakoudakis2011new, persing2013modeling} to neural, trait-specific models~\cite{taghipour2016neural, uto2020neural}, and recently to LLM-based evaluation~\cite{lee2024unleashing, chu-etal-2025-rationale}. These typically score coarse-grained aspects like grammar, coherence, content, and creativity~\cite{li2024automated}, but remain English-centric and overlook the rhetorical and cultural complexities of Chinese writing. In addition, although recent frameworks like \textit{WritingBench}~\cite{wu2025writingbench} and \textit{BigGen Bench}~\cite{kim-etal-2025-biggen} offer fine-grained evaluation strategies through prompt-specific assessment instances, they fall short in covering a broader range of writing prompts, limiting their applicability to various essay tasks.

\section{Conclusion}
In this work, we present \benchName, the first comprehensive benchmark for evaluating the capabilities of LLMs in the Chinese essay writing and evaluation across four distinct literary genres. To address the challenges of analytic and accurate essay evaluation, \benchName adopts a genre-oriented, hierarchical multi-trait evaluation approach that enables fine-grained scoring. Specifically, we introduce a dependency-based aggregation strategy to compute the final scores. Our comprehensive human agreement study and sensitivity analysis demonstrate that the framework achieves high alignment with human judgment and effectively distinguishes essays of varying quality. Furthermore, we benchmark 15 large-size LLMs on Chinese essay writing, revealing notable limitations in descriptive and narrative essays, particularly for open-ended prompts. Overall, \benchName offers a diverse dataset and a robust evaluation framework for Chinese essay, with practical implications for educational applications and future research.

% \begin{table*}
% \centering
% \begin{tabular}{llc}
% \hline
% \textbf{Instruction} & \textbf{Example} & \textbf{Evaluation}\\
% \hline
% Genre & "Write an argumentative writing about ..." & Rule \\
% Topic & "Write an essay about winter" & LLM \\
% Words & "no more than 1500 words; over 1200 words" & Rule\\
% Content & "The writing should include the weather description " & LLM \\
% Readings & "Read the following story and write an essay" & Rule \\
% Reader & "The oriented reader is the youth" & Rule \\
% \hline
% \end{tabular}
% \caption{Example commands for accented characters, to be used in, \emph{e.g.}, Bib\TeX{} entries.}
% \label{tab:accents}
% \end{table*}

\section*{Limitations}
Despite the contributions presented in our work, several minor limitations remain:
\begin{itemize}
    \item First, the datasets and evaluation dimensions used in this study are primarily based on Chinese essay prompts. While the widely adopted essay categorization framework can be applied to other languages such as English and Japanese, the current work focuses on Chinese essay writing. This is due to the significant differences in idioms, linguistic conventions, and cultural expressions across languages. Nonetheless, the framework has the potential to be translated and extended to multilingual settings in future work.
    
    \item Second, although this work proposes a more fine-grained evaluation method for Chinese essays, the designed evaluation traits primarily focus on overall expression and structural aspects, such as paragraph organization and comprehensive performance of the essays from multiple perspectives. However, it overlooks more granular analyses at the lexical and sentence levels. Future research could incorporate finer-grained evaluations that consider sentence-level coherence and word-level richness
    \item Third, \benchName primarily focuses on evaluating the overall quality of essays, while overlooking instruction-following abilities. For example, whether the generated essays adhere strictly to the prompt requirements has not been thoroughly assessed. To enable a more comprehensive evaluation, future research could address this gap by incorporating the essay prompt following ability as an explicit evaluation dimension.
\end{itemize}

\section*{Ethics Statement}
To mitigate potential ethical concerns, all essay prompts were carefully reviewed and filtered by manual inspection. We ensured that none of the prompts contained offensive, gender-biased, harmful, or otherwise ethically inappropriate content. In addition, all participants involved in the human agreement study were professional annotators who were fairly compensated for their contributions.

% \section*{Ethics Statement}
% Scientific work published at ACL 2023 must comply with the ACL Ethics Policy.\footnote{\url{https://www.aclweb.org/portal/content/acl-code-ethics}} We encourage all authors to include an explicit ethics statement on the broader impact of the work, or other ethical considerations after the conclusion but before the references. The ethics statement will not count toward the page limit (8 pages for long, 4 pages for short papers).

\bibliography{anthology,custom}
\bibliographystyle{acl_natbib}

\clearpage

% \appendix

% \section{Appendix}
% \label{sec:appendix}

\appendix
% \label{sec:appendix}

\section{Performance across different genres}
\begin{figure}
    \centering
    \includegraphics[width=\linewidth]{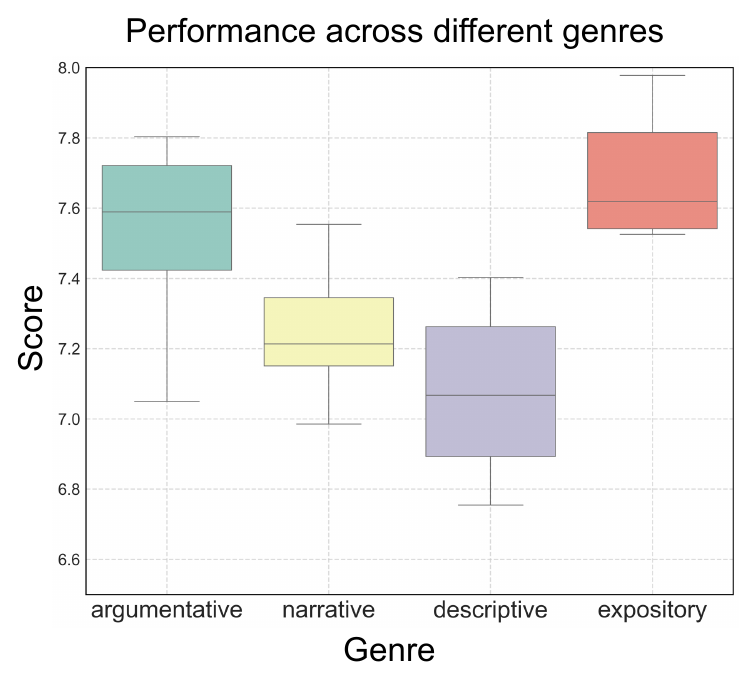}
    \caption{Comparison of LLMs Performance across different genres.}
    \label{fig:genre_performance}
\end{figure}

\section{Hierarchical Traits Design}
\label{appendix:traits}

We developed a comprehensive, genre-specific multi-trait evaluation framework tailored explicitly to the distinctive features, objectives, and contexts of each essay genre. This system aligns closely with educational standards and requirements, ensuring a precise and relevant assessment approach.

\subsection{Argumentative Essays}
For Argumentative essays, we focus on:
\begin{itemize}
\item \textbf{Argument} ($d=0$): Clarity, precision, and relevance of the central viewpoint.
\item \textbf{Evidence} ($d=1$): Strength, appropriateness, and reliability of supporting details and examples.
\item \textbf{Argumentation Methods} ($d=2$): Effective use of logical strategies such as comparison, causality, and deduction.
\item \textbf{Logical Development} ($d=2$): Coherent and logically sequenced progression of ideas.
\item \textbf{Expression} ($d=3$): Clarity, precision, and stylistic appropriateness of language.
\item \textbf{Endings} ($d=3$): Ability to summarize arguments effectively and deliver a compelling conclusion.
\end{itemize}

\subsection{Narrative Essays}
For Narrative essays, we emphasize:
\begin{itemize}
\item \textbf{Language and Style} ($d=0$): Effectiveness of vocabulary, tone, and stylistic choices in storytelling.
\item \textbf{Structural Layer and Narrative Techniques} ($d=0$): Sophisticated use of narrative structures and literary techniques.
\item \textbf{Theme and Emotional Expression} ($d=1$): Depth of thematic content and authenticity of emotional portrayal.
\item \textbf{Overall Structure and Plot Development} ($d=2$): Clear progression, effective pacing, and cohesive plot structure.
\item \textbf{Characterization and Detail} ($d=2$): Rich, vivid portrayal of characters and setting details.
\item \textbf{Choice of Material} ($d=3$): Originality, relevance, and effectiveness in selecting narrative content.
\end{itemize}

\subsection{Descriptive Essays}
For Descriptive essays, we highlight:
\begin{itemize}
\item \textbf{Clarity of Subject and Central Theme} ($d=0$): Distinct and clearly communicated central image or idea.
\item \textbf{Rhythm and Overall Fluency} ($d=1$): Smooth flow and harmonious pacing throughout the essay.
\item \textbf{Content and Unique Perspective} ($d=1$): Original insights and distinctive angles in the descriptions.
\item \textbf{Structure and Organization} ($d=2$): Effective and logical arrangement enhancing readability.
\item \textbf{Emotional Expression and Atmosphere Description} ($d=2$): Authentic depiction of the atmosphere and emotional tone.
\item \textbf{Sensory Details} ($d=3$): Use of vivid and engaging sensory imagery.
\end{itemize}

\subsection{Expository Essays}

Finally, for Expository essays, we prioritize:
\begin{itemize}
\item \textbf{Clarity of Topic and Purpose} ($d=0$): Clearly defined subject matter and objectives.
\item \textbf{Practicality and Relevance} ($d=0$): Real-world applicability and pertinence of the provided information.
\item \textbf{Scientific Accuracy and Credibility of Content} ($d=0$): Validity and trustworthiness of the facts and data presented.
\item \textbf{Logical Structure and Coherence} ($d=1$): Systematic and logically sound organization of ideas.
\item \textbf{Clarity and Appropriateness of Language} ($d=2$): Use of a clear, accessible, and appropriate academic language.
\item \textbf{Diversity and Appropriateness of Explanatory Methods} ($d=3$): Variety and suitability of explanatory techniques, enhancing comprehension and reader engagement.
\end{itemize}

% We design the multi-traits for evaluating different genre's essays based on their unique features, aims, and usage scenarios, referring to the educational requirements. In details, the Argumentative traits are:
% \begin{itemize}
%     \item Argument ($d=0$)
%     \item Evidence ($d=1$)
%     \item Argumentation Methods ($d=2$)
%     \item Logical Development ($d=2$)
%     \item Expression ($d=3$)
%     \item Endings ($d=3$)
% \end{itemize}

% the Narrative traits are:
% \begin{itemize}
%     \item Language and Style ($d=0$)
%     \item Structural Layer and Narrative Techniques ($d=0$)
%     \item Theme and Emotional Expression ($d=1$)
%     \item Overall Structure and Plot Development ($d=2$)
%     \item Characterization and Detail ($d=2$)
%     \item Choice of Material ($d=3$)
% \end{itemize}

% the Descriptive Traits are:
% \begin{itemize}
%     \item Clarity of Subject and Central Theme ($d=0$)
%     \item Rhythm and Overall Fluency ($d=1$)
%     \item Content and Unique Perspective ($d=1$)
%     \item Structure and Organization ($d=2$)
%     \item Emotional Expression and Atmosphere Description ($d=2$)
%     \itemize Sensory Details ($d=3$)
% \end{itemize}
% The expository essays are
% \begin{itemize}
%     \item Clarity of Topic and Purpose ($d=0$)
%     \item Practicality and Relevance ($d=0$)
%     \item Scientific Accuracy and Credibility of Content ($d=0$)
%     \item Logical Structure and Coherence ($d=1$)
%     \item Clarity and Appropriateness of Language ($d=2$)
%     \item Diversity and Appropriateness of Explandatory Methods ($d=3$)
% \end{itemize}

\section{Evaluation Prompt}
\label{appendix: prompts}
\subsection{CoT Prompting}
\begin{figure*}
\small
    \centering
    \includegraphics[width=0.75\textwidth]{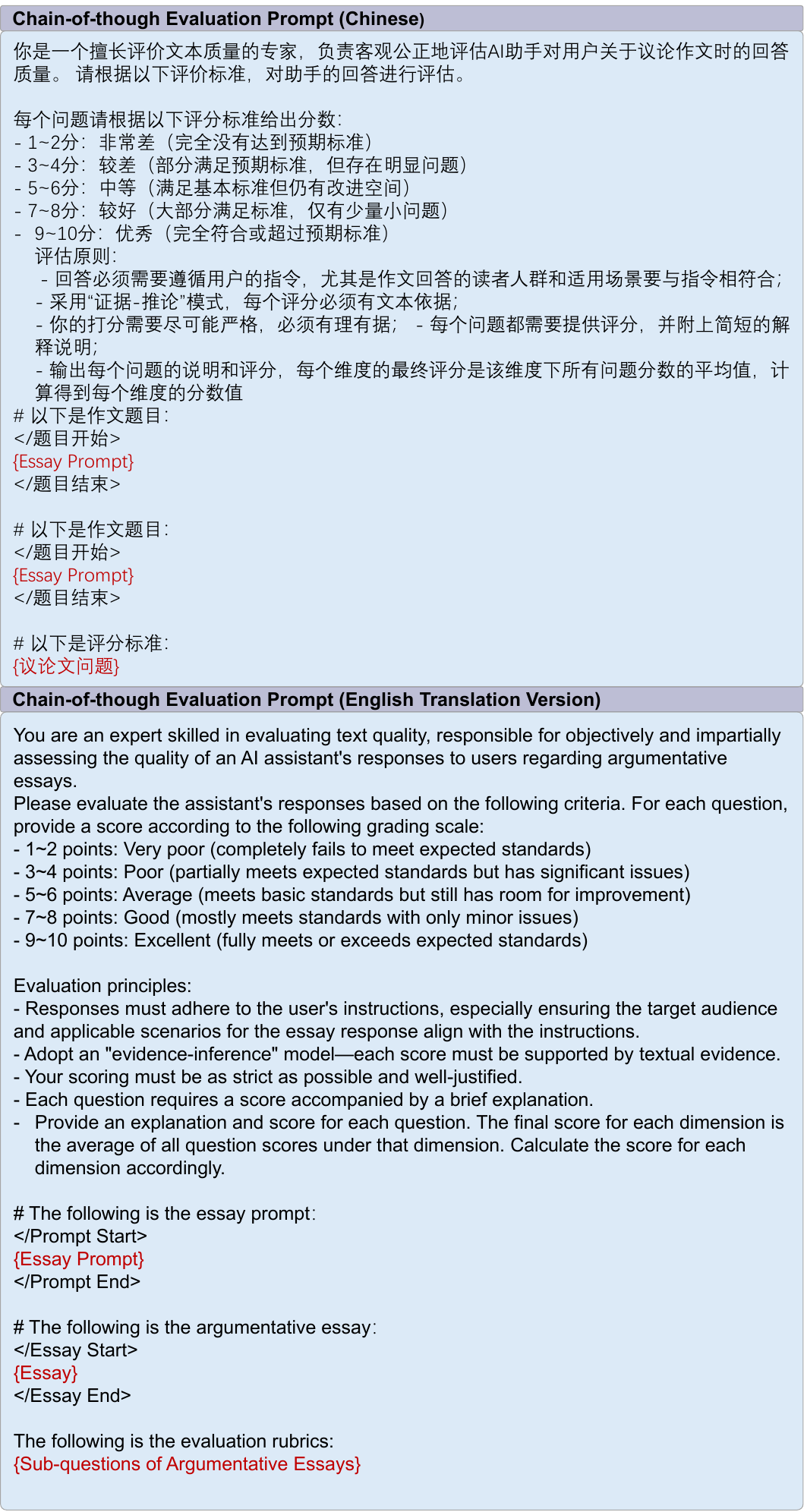}
    \caption{CoT Prompt Strategy for Evaluation.}
    \label{fig:general_prompt}
\end{figure*}

In our evaluation method, we implement the Chain-of-Thought (CoT) prompting strategy, which first guides the large language models (LLMs) to systematically analyze essays before assigning scores. This structured analytical step provides LLMs with robust reasoning and clear justifications, facilitating accurate scoring decisions. In addition, the detailed CoT reasoning process serves as a valuable reference, allowing evaluators and users to better understand and verify the rationale behind each assigned score. The specific prompts used for the CoT strategy are illustrated in Figure~\ref{fig:general_prompt}.
\subsection{Trait-based sub-questions}
\begin{figure*}
\small
    \centering
    \includegraphics[width=\linewidth]{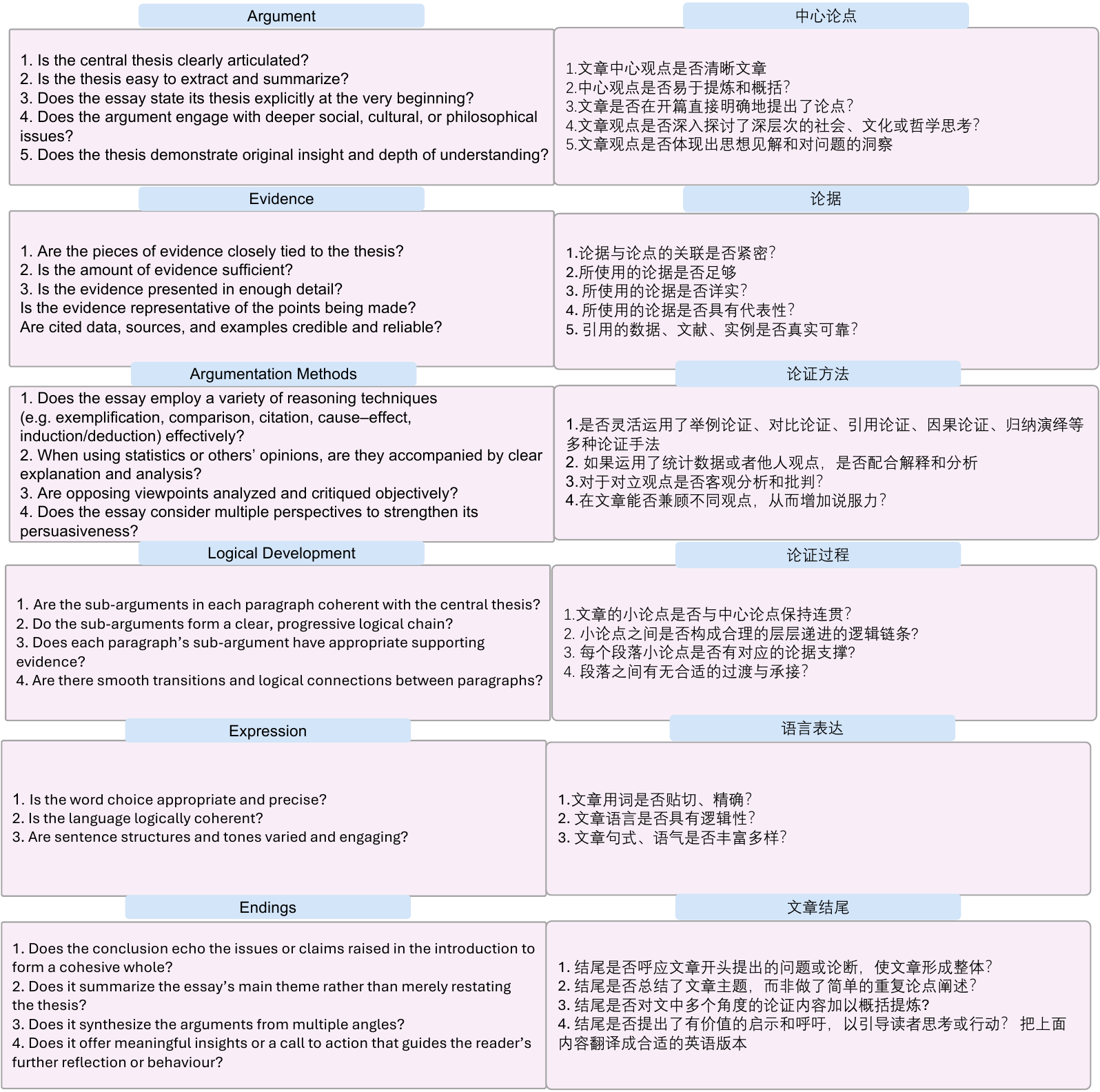}
    \caption{Multi-traits and sub-questions of Argumentative Essay.}
    \label{fig:argumentative}
\end{figure*}

\begin{figure*}
\small
    \centering
    \includegraphics[width=\linewidth]{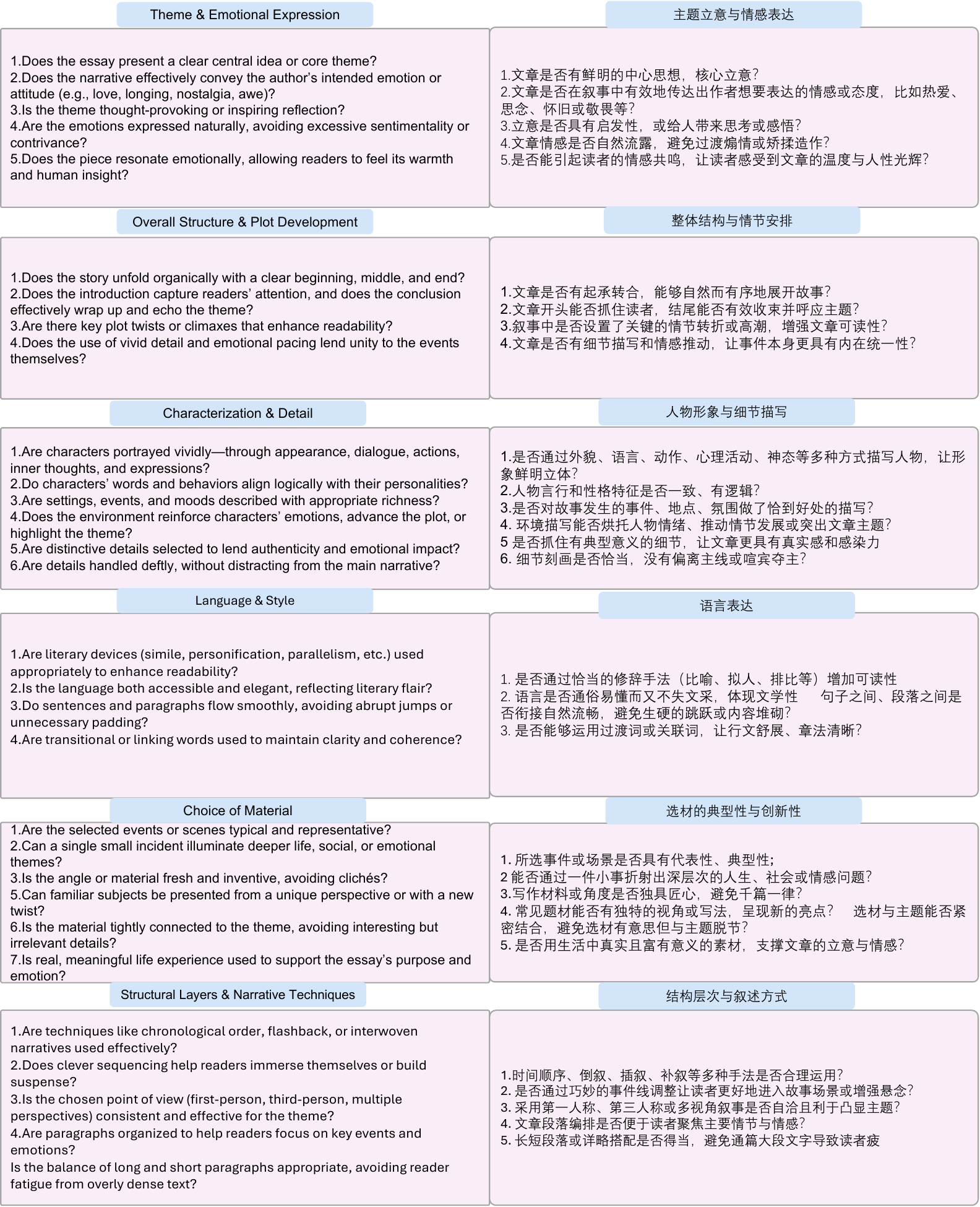}
    \caption{Multi-traits and sub-questions of Narrative Essay.}
    \label{fig:narrative}
\end{figure*}

\begin{figure*}
\small
    \centering
    \includegraphics[width=\linewidth]{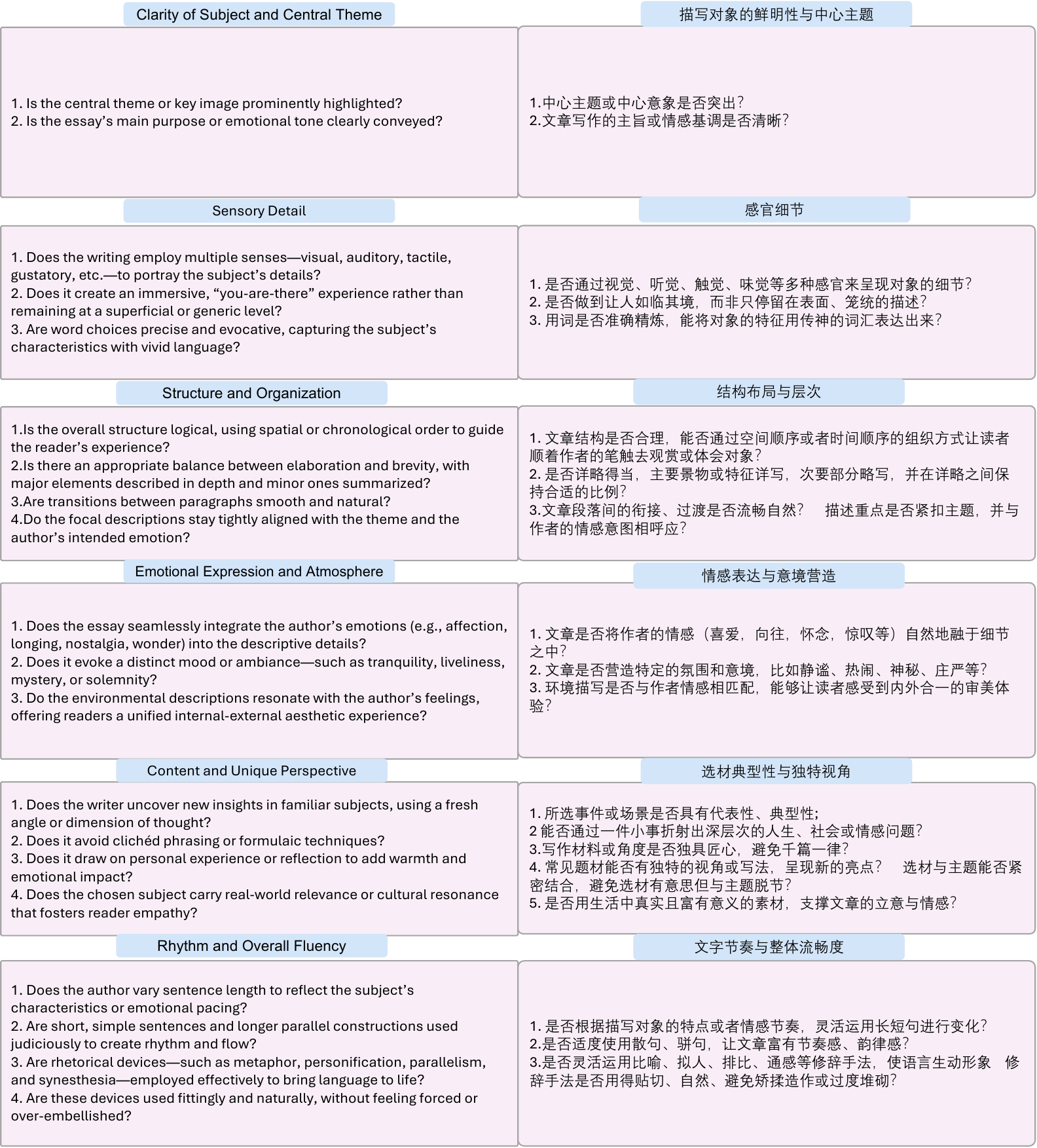}
    \caption{Multi-traits and sub-questions of Descriptive Essay.}
    \label{fig:descriptive}
\end{figure*}

\begin{figure*}
\small
    \centering
    \includegraphics[width=\linewidth]{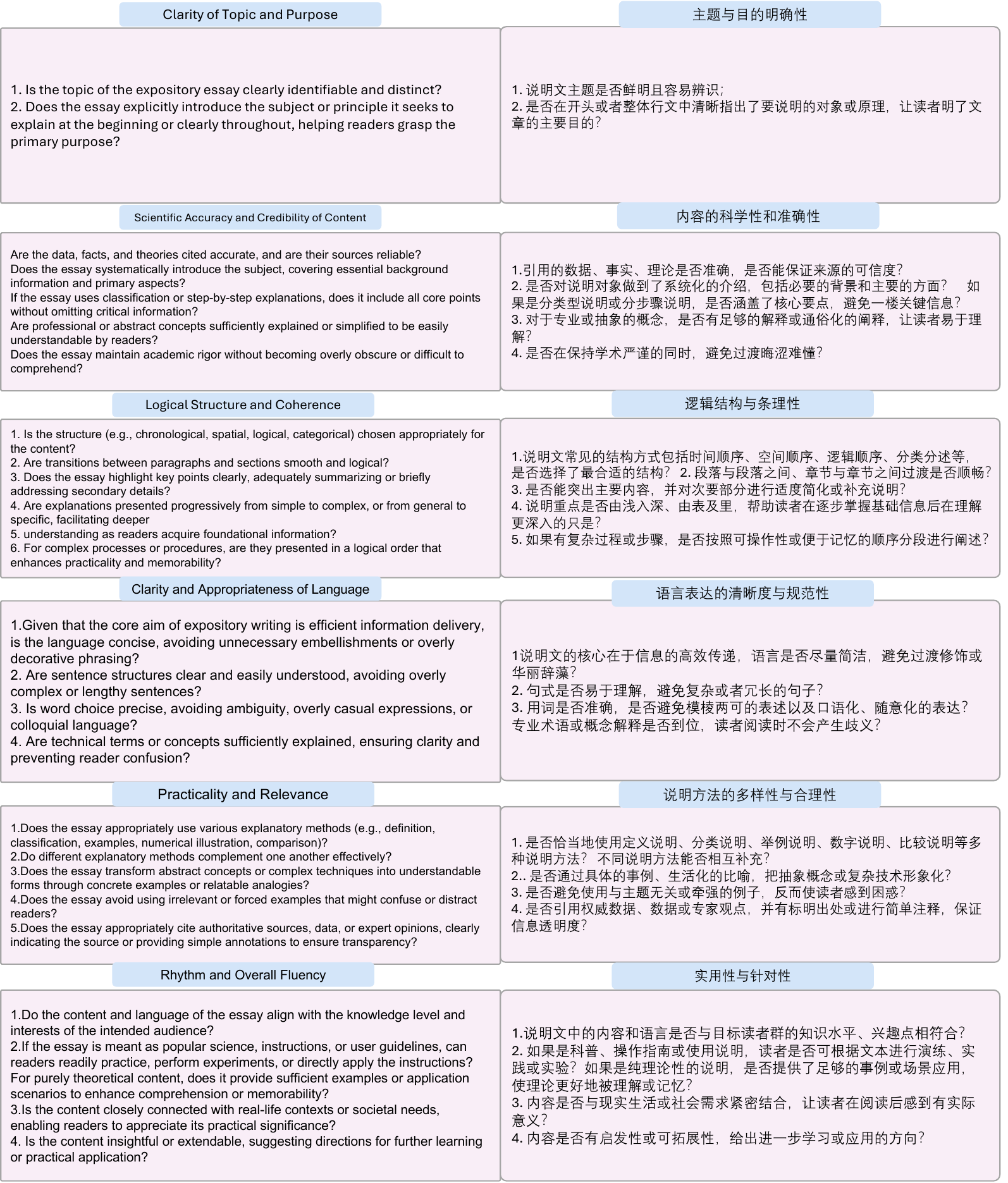}
    \caption{Multi-traits and sub-questions of Expository Essay.}
    \label{fig:expository}
\end{figure*}

For each hierarchically designed trait, we carefully develop a series of detailed, targeted evaluation questions, addressing multiple dimensions and perspectives relevant to each trait. These questions are crafted to comprehensively assess the specific characteristics and nuances inherent in each genre of essays. The specific questions tailored for Argumentative, Narrative, Descriptive, and Expository essays are illustrated in Figures~\ref{fig:argumentative}, \ref{fig:narrative}, \ref{fig:descriptive}, and \ref{fig:expository}, respectively.

\section{Human Annotation}

\begin{figure*}
\small
    \centering
    \includegraphics[width=0.7\linewidth]{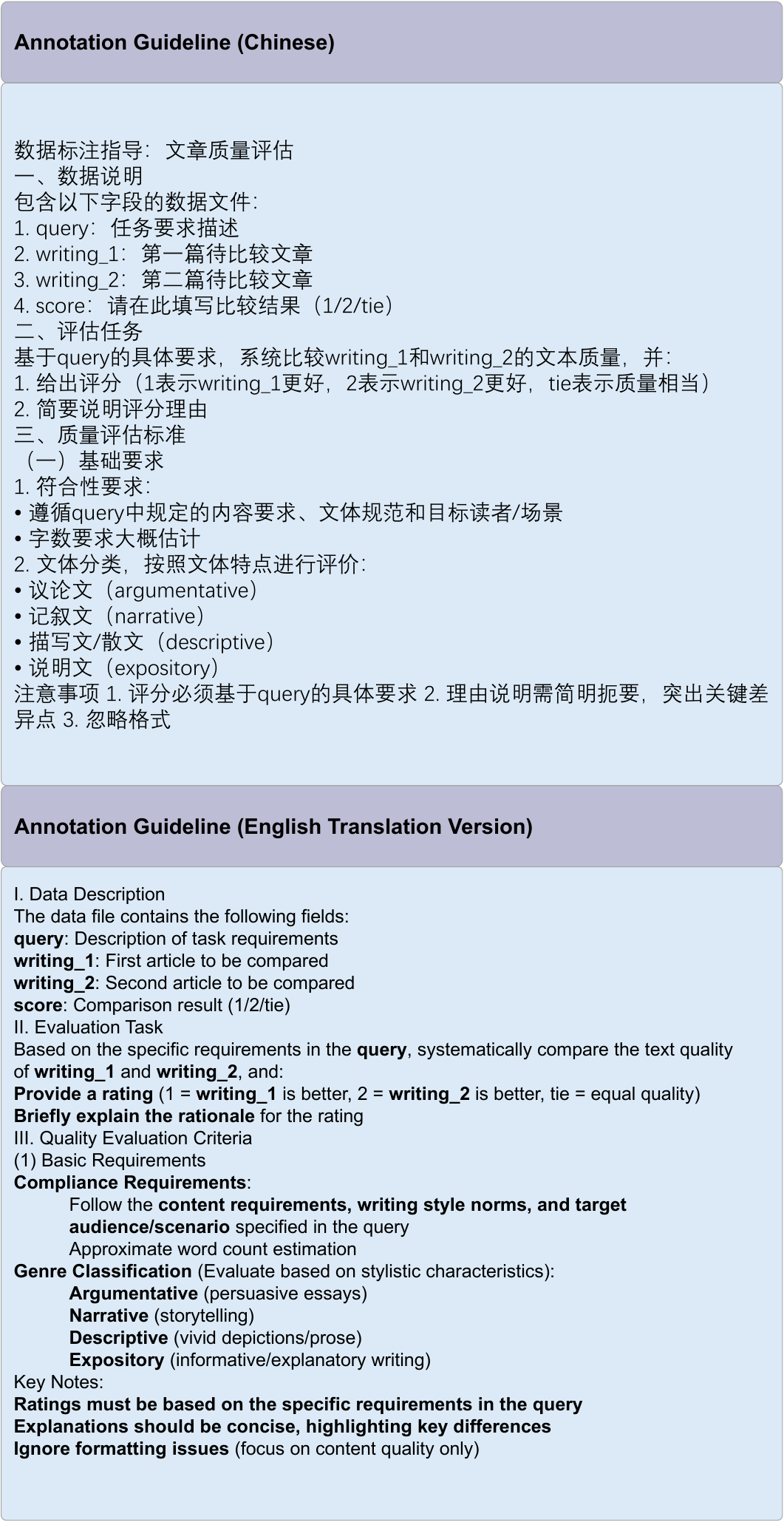}
    \caption{Annotation Guideline for Human Agreement Study.}
    \label{fig:annotation}
\end{figure*}

For our human agreement study, we recruited 14 professional annotators with strong backgrounds in Chinese linguistics and language education. Each annotator was assigned approximately 70 data samples per day, working within an 8-hour schedule. The complete annotation of 5,040 data items was completed over five days. This rigorous process ensured consistency, reliability, and high-quality annotations across the dataset. The comprehensive annotation guidelines provided to annotators are illustrated in Figure~\ref{fig:annotation}.

\section{Case Study}

\begin{figure*}
\small
    \centering
    \includegraphics[width=\linewidth]{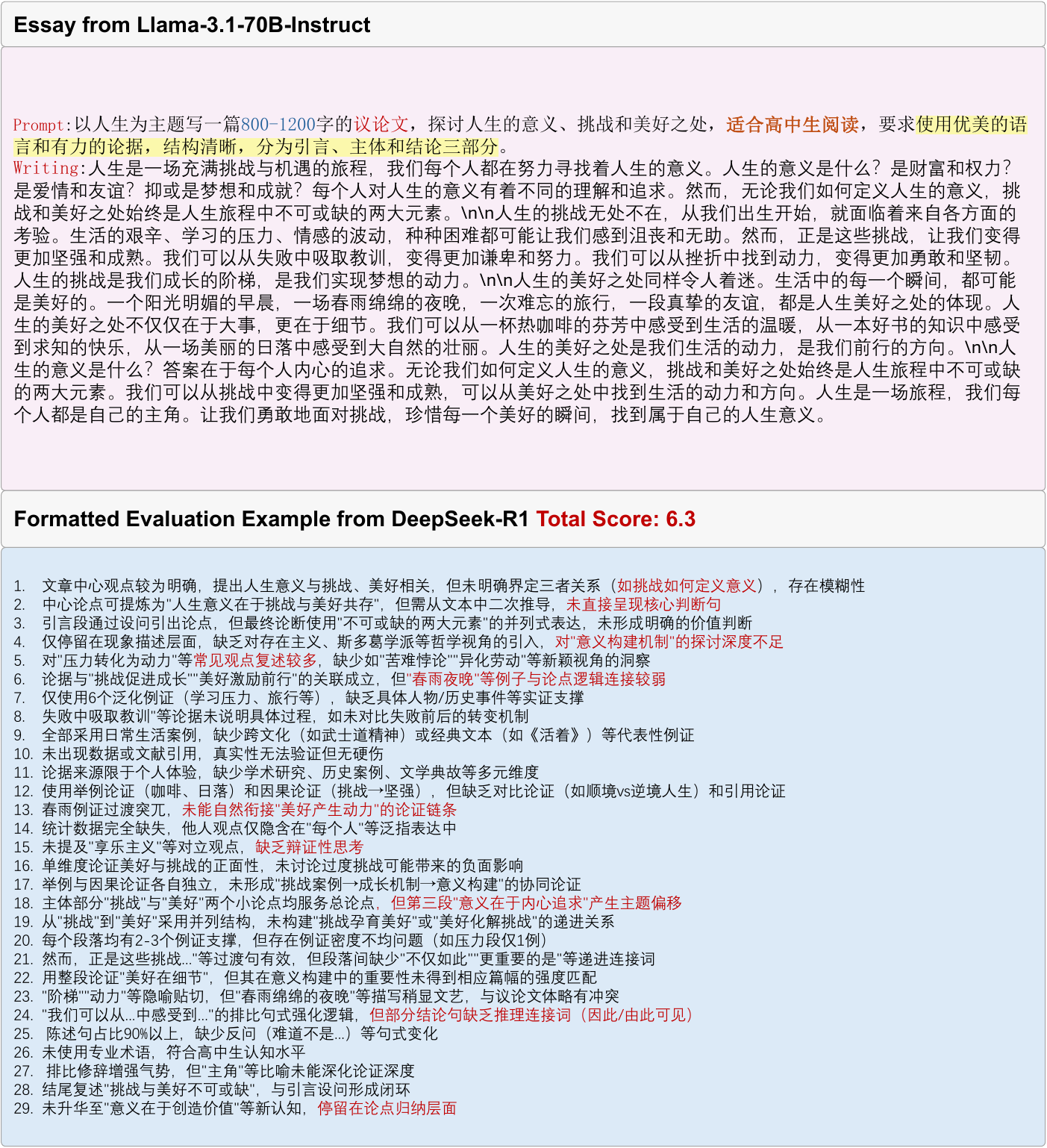}
    \caption{Low-quality Essay and Evaluation Example.}
    \label{fig:low_quality}
\end{figure*}

\begin{figure*}
\small
    \centering
    \includegraphics[width=\linewidth]{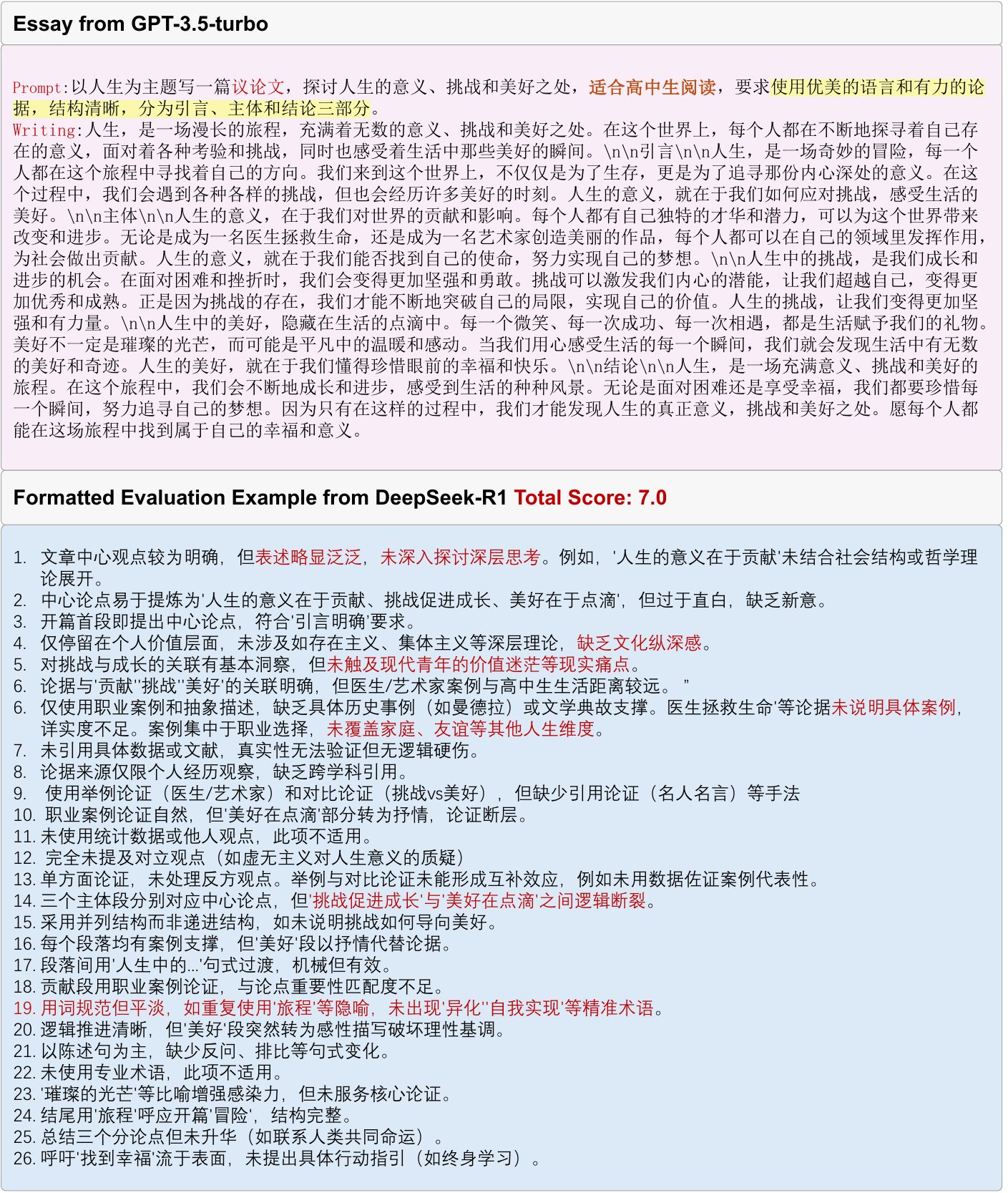}
    \caption{Median-quality Essay and Evaluation Example.}
    \label{fig:median_quality}
\end{figure*}

\begin{figure*}
\small
    \centering
    \includegraphics[width=\linewidth]{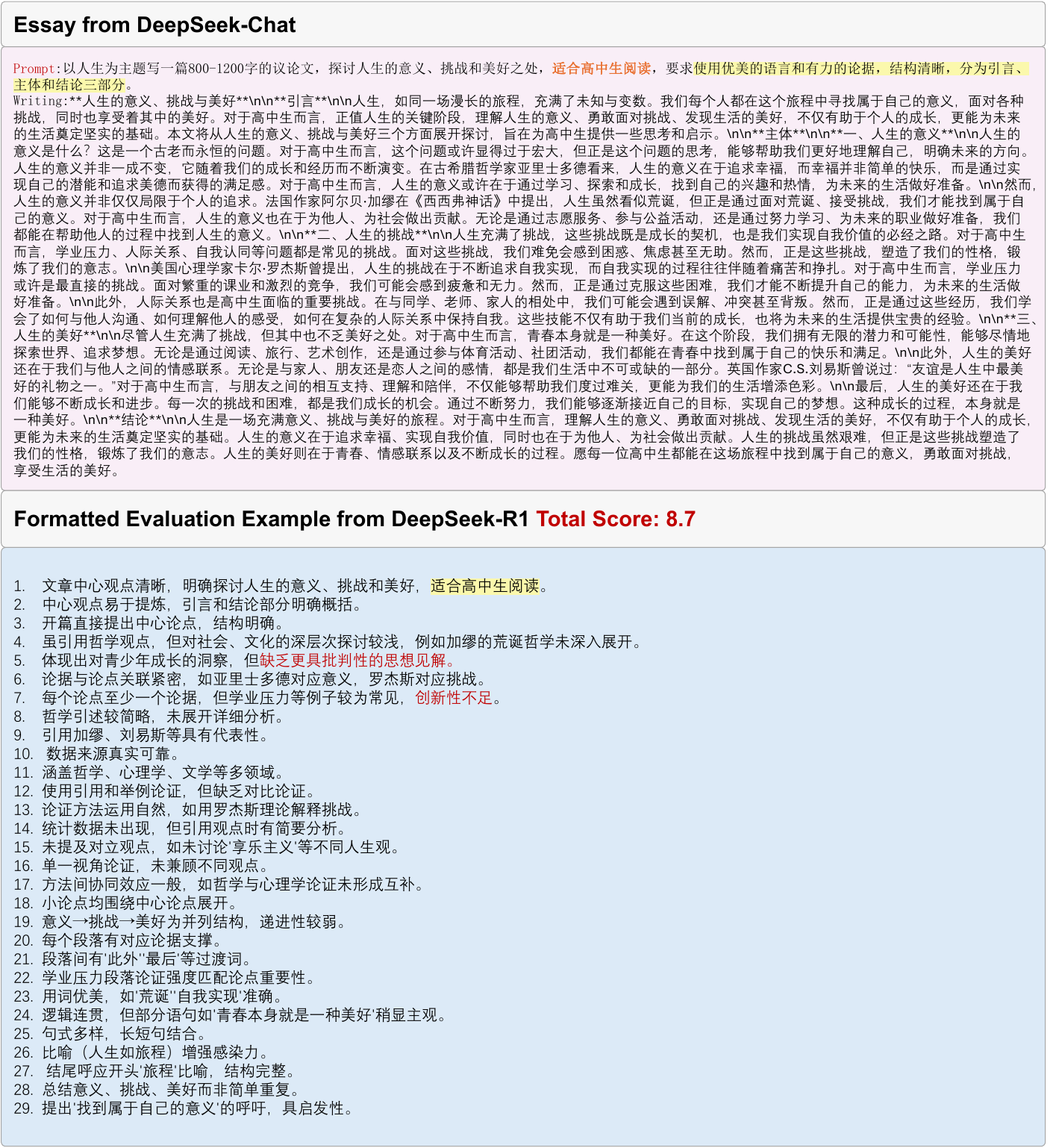}
    \caption{High-quality Essay and Evaluation Example.}
    \label{fig:high_quality}
\end{figure*}

In this section, we conduct a qualitative case study of essays across different quality levels, using outputs from various LLMs. Specifically, we examine high-quality essays generated by DeepSeek Chat (score: $8.7$), median-quality essays by GPT-3.5-turbo (score: $7.0$), and low-quality essays by Llama-3.1-70B-Instruct (score: $6.3$), as shown in Figures~\ref{fig:low_quality}, \ref{fig:median_quality}, and \ref{fig:high_quality}. These scores are derived from the evaluations conducted by DeepSeek-R1. We observe that the evaluator provides detailed and consistent analyses across essays of varying quality, highlighting both strengths and weaknesses. This type of evaluative feedback demonstrates strong interpretability and reliability, making it valuable for future educational applications such as formative writing assessment and personalized feedback generation.

% This is a section in the appendix.

\end{document}